%% file: sample-sigconf.tex
\documentclass[acmtog]{acmart}
\usepackage{subcaption}
\usepackage{booktabs}
\usepackage{colortbl}
\usepackage{xcolor}
\usepackage{multirow}

\definecolor{bestColor}{HTML}{FFB74D}   
\definecolor{secondColor}{HTML}{FFCC80} 
\definecolor{thirdColor}{HTML}{FFE0B2}

\newcommand{\best}[1]{\cellcolor{bestColor}#1}
\newcommand{\second}[1]{\cellcolor{secondColor}#1}
\newcommand{\third}[1]{\cellcolor{thirdColor}#1}
\AtBeginDocument{%
  }

\setcopyright{cc}
\setcctype{by}
\copyrightyear{2026}
\acmJournal{TOG}
\acmYear{2026} \acmVolume{45} \acmNumber{4} \acmArticle{51}
\acmMonth{7} \acmDOI{10.1145/3811304}
\acmConference[SIGGRAPH '26]{Special Interest Group on Computer Graphics and Interactive Techniques Conference}{July 19--23, 2026}{Los Angeles, CA, USA}


\acmSubmissionID{320}

\usepackage[capitalize]{cleveref}
\crefname{section}{Sec.}{Secs.}
\Crefname{section}{Section}{Sections}
\Crefname{table}{Table}{Tables}
\crefname{table}{Tab.}{Tabs.}
\Crefname{algorithm}{Algorithm}{Algorithms}
\crefname{algorithm}{Alg.}{Algs.}

\usepackage{xspace}
\makeatletter
\DeclareRobustCommand\onedot{\futurelet\@let@token\@onedot}
\def\@onedot{\ifx\@let@token.\else.\null\fi\xspace}

\makeatother

\defcitealias{layerdiffuse}{Zhang et al.}
\begin{document}


\title{UniVidX: A Unified Multimodal Framework for Versatile Video Generation via Diffusion Priors}
\author{Houyuan Chen}
\orcid{0009-0005-4693-2326}
\email{houyuanchen111@gmail.com}
\affiliation{%
  \institution{MMLab@HKUST}
  \city{Hong Kong}
  \country{China}
}

\author{Hong Li}
\orcid{0000-0002-4240-3073}
\email{link0502@buaa.edu.cn}
\affiliation{%
  \institution{Beihang University}
  \city{Beijing}
  \country{China}
}

\author{Xianghao Kong}
\orcid{0009-0004-9865-4105}
\email{refkxh@outlook.com}
\affiliation{%
  \institution{MMLab@HKUST}
  \city{Hong Kong}
  \country{China}
}

\author{Tianrui Zhu}
\orcid{0009-0002-5146-019X}
\email{221900034@smail.nju.edu.cn}
\affiliation{%
  \institution{Nanjing University}
  \city{Nanjing}
  \country{China}
}

\author{Shaocong Xu}
\orcid{0000-0001-7525-0790}
\email{scxu@baai.ac.cn}
\affiliation{%
  \institution{BAAI}
  \city{Beijing}
  \country{China}
}

\author{Weiqing Xiao}
\orcid{0009-0003-2548-0485}
\email{weiqing001@smail.nju.edu.cn}
\affiliation{%
  \institution{Nanjing University}
  \city{Nanjing}
  \country{China}
}

\author{Yuwei Guo}
\orcid{0009-0003-1516-4083}
\email{guoyw.nju@gmail.com}
\affiliation{%
  \institution{MMLab@CUHK}
  \city{Hong Kong}
  \country{China}
}

\author{Chongjie Ye}
\orcid{0000-0002-7123-0220}
\email{chongjieye@link.cuhk.edu.cn}
\affiliation{%
  \institution{CUHK-Shenzhen}
  \city{Shenzhen}
  \country{China}
}

\author{Lvmin Zhang}
\orcid{0000-0003-3503-5791}
\email{lyuminzhang@outlook.com}
\affiliation{%
  \institution{Stanford University}
  \city{Stanford}
  \country{USA}
}

\author{Hao Zhao}
\orcid{0000-0001-7903-581X}
\email{zhaohao@air.tsinghua.edu.cn}
\affiliation{%
  \institution{Tsinghua University}
  \city{Beijing}
  \country{China}
}

\author{Anyi Rao}
\authornote{Corresponding author.}
\orcid{0000-0003-1004-7753}
\email{anyirao@ust.hk}
\affiliation{%
  \institution{MMLab@HKUST}
  \city{Hong Kong}
  \country{China}
}

\renewcommand{\shortauthors}{Chen et al.}

\begin{abstract}
Recent progress has shown that video diffusion models (VDMs) can be repurposed to solve various multimodal graphics tasks.
However, existing approaches predominantly train separate models for each specific problem setting. 
This practice locks models into fixed input-output mappings, and typically ignores the joint correlations across modalities.
In this paper, we present \textbf{UniVidX}, a unified multimodal framework designed to leverage VDM priors to enable versatile video generation. 
Our goal is to~(i) master diverse pixel-aligned tasks by formulating them as conditional generation problems within multimodal space,~(ii) adapt to modality-specific distributions without compromising the backbone’s native priors, and~(iii) ensure cross-modal consistency during synthesis.
Concretely, we propose three key designs:~1) Stochastic Condition Masking (SCM): by randomly partitioning modalities into clean conditions and noisy targets during training, we enable the model to learn omni-directional conditional generation rather than fixed mappings.~2) Decoupled Gated LoRA (DGL): we attach per-modality LoRAs and activate them when a modality serves as a generation target, thereby preserving the VDM's strong priors.~3) Cross-Modal Self-Attention (CMSA): we explicitly share keys/values across modalities while maintaining modality-specific queries, facilitating information exchange and inter-modal alignment.
We validate our framework by instantiating it in two domains:~1) \textit{UniVid-Intrinsic} for RGB videos and their intrinsic maps (albedo, irradiance, normal), and~2) \textit{UniVid-Alpha} for blended RGB videos and their constituent RGBA layers. 
Experimental results demonstrate that both models achieve performance competitive with state-of-the-art methods across distinct tasks. 
Notably, they exhibit robust generalization capabilities in in-the-wild scenarios, even when trained on limited datasets of fewer than 1k videos. 
Our project page: \noindent\textcolor{blue}{\url{https://houyuanchen111.github.io/UniVidX.github.io/}}.
\end{abstract}

%



\maketitle

\section{Introduction}
\begin{figure*}[t]
  \centering
  \includegraphics[width=\textwidth]{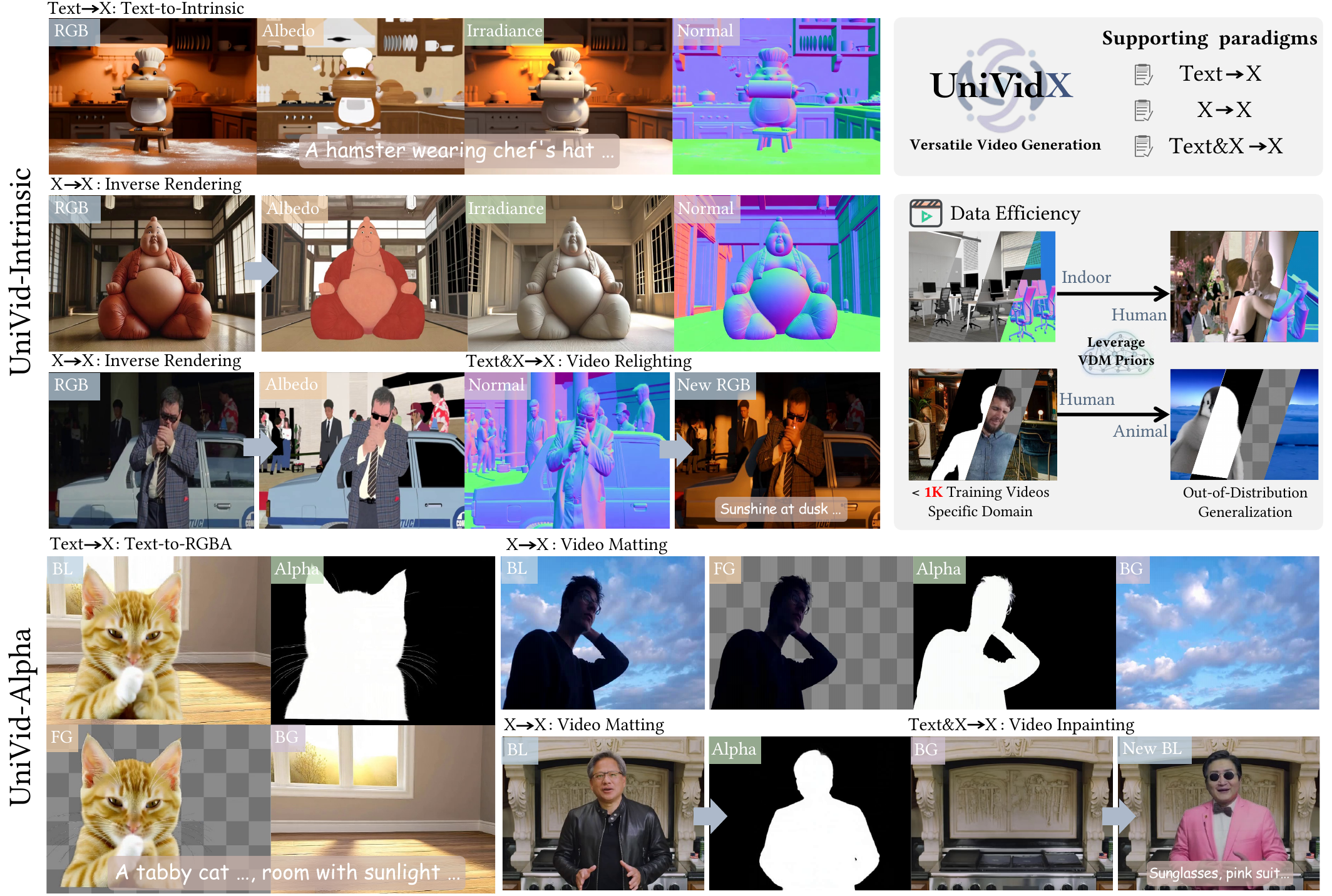}
  \caption{
\textbf{UniVidX} is a unified multimodal framework designed for versatile video generation, which supports diverse paradigms (Text$\to$X, X$\to$X, and Text\&X$\to$X; 'X' denotes visual modality like albedo).
We instantiate this framework into two models: 1) \textit{UniVid-Intrinsic} (top), which supports tasks including text-to-intrinsic, inverse rendering, and video relighting; and 2) \textit{UniVid-Alpha} (bottom), which supports tasks including text-to-RGBA, video matting, and video inpainting. 
Notably, by leveraging VDM priors, both models demonstrate remarkable data efficiency, generalizing well despite being trained with small-scale data.
}
  \label{fig:teaser}
\end{figure*}

Pre-trained Video Diffusion Models (VDMs) have evolved into powerful foundation engines, capturing rich priors of real-world dynamics~\cite{stablevideodiffusion,sora,opensora,opensora2,cogvideo,cogvideox,hunyuanvideo,wan}. 
Leveraging the robust VDM priors for downstream multimodal graphics tasks, ranging from perception (e.g., intrinsic decomposition~\cite{diffusionrenderer}) to generation (e.g., content creation~\cite{wanalpha}), has proven to be highly effective.

However, existing approaches typically treat different problems in isolation, training separate networks for each specific input–output mapping (e.g., RGB$\to$alpha; intrinsic$\to$X), which introduces two critical limitations.  
First, it locks each model into a fixed role, limiting flexibility for diverse graphics applications where input conditions may vary.  
Second, it often ignores the correlations shared across visual modalities~\cite{taskonomy,omnidata}, an oversight reflected in their modality-exclusive prediction strategy. 
This restricts prior methods to either dedicated single-modality generation (e.g., NormalCrafter~\cite{normalcrafter}) or serial multimodal inference (e.g., Ouroboros~\cite{ouroboros}), which leads to cross-modal inconsistencies in the final modality stack.

Motivated by this limitation, we pose a fundamental question:
\textit{
Can we design a unified generative framework that allows a video model to let different subsets of aligned modalities set act as conditions or targets}, enabling flexible generation across visual modalities?

Realizing such a unified formulation is non-trivial and presents three primary challenges:  
(i) It must be capable of mastering diverse task categories within a single conditional generation framework;  
(ii) It requires adapting to distinct modality distributions, while simultaneously preserving the backbone's generative priors to ensure high-quality output;  
and (iii) It must guarantee alignment across diverse interacting modalities during joint generation.

To this end, we present \textbf{UniVidX}. 
It is a unified multimodal framework designed to leverage VDM priors for versatile video generation, which incorporates three key designs: \textbf{1)~Stochastic Condition Masking~(SCM)} randomly partitions modalities into clean conditions and noisy targets, enabling the T2V backbone to uniformly process pure text, visual, and hybrid inputs, thereby compelling the model to learn omni-directional generation. 
\textbf{2)~Decoupled Gated LoRA~(DGL)} assigns independent LoRAs~\cite{lora} to each modality and activates them only when that modality is a generation target, preventing parameter interference while preserving VDM priors; 
and \textbf{3)~Cross-Modal Self-Attention~(CMSA)}, where keys and values are shared across modalities while queries remain modality-specific to ensure cross-modal consistency.

To validate the effectiveness of our framework, we instantiate \textbf{UniVidX} in two multimodal domains: 1) \textit{UniVid-Intrinsic}, which models among RGB videos and the corresponding intrinsic maps (albedo/irradiance/normal), and 2) \textit{UniVid-Alpha}, which processes blended RGB (BL), alpha matte (Alpha), foreground (FG), and background (BG) layers. 
Powered by unified design of our \textbf{UniVidX}, both models demonstrate versatility, supporting three paradigms (Text$\to$X; X$\to$X; Text\&X$\to$X) and collectively covering \textbf{15} distinct tasks. 
As illustrated in Fig.~\ref{fig:teaser}, \textit{UniVid-Intrinsic} (top) can handle tasks such as text-to-intrinsic (Text$\to$X), inverse rendering (X$\to$X), and video relighting (Text\&X$\to$X);
\textit{UniVid-Alpha} (bottom) enables tasks including text-to-RGBA (Text$\to$X), video matting (X$\to$X), and video inpainting (Text\&X~$\to$X). 
Moreover, the flexibility of our approach allows for the composition of different tasks to support downstream applications, such as video relighting, video retexturing, material editing for \textit{UniVid-Intrinsic}, and video inpainting, background/foreground replacement for \textit{UniVid-Alpha} (see Sec.~\ref{sec:application}).

Remarkably, attributed to the efficient utilization of VDM priors, both models demonstrate exceptional data efficiency. 
They exhibit robust generalization to out-of-distribution, in-the-wild scenarios, despite being trained on limited domain-specific datasets. 
Moreover, extensive experiments demonstrate that both \textit{UniVid-Intrinsic} and \textit{UniVid-Alpha} achieve performance competitive with state-of-the-art methods across diverse tasks. 
The main contributions of this work are summarized as follows:
1)~We propose \textbf{UniVidX}, a unified multimodal framework that utilizes video diffusion priors to enable versatile generation across diverse visual modalities.
2)~We introduce Stochastic Condition Masking~(SCM) for omni-directional generation, Decoupled Gated LoRA (DGL) for preventing parameter interference and preserving native priors, and Cross-Modal Self-Attention (CMSA) for cross-modal consistency.
3)~We validate our framework by instantiating it into two distinct models, \textit{UniVid-Intrinsic} and \textit{UniVid-Alpha}. Both demonstrate state-of-the-art performance across diverse tasks and robust in-the-wild generalization, despite using limited training data ($<$1k videos).

\section{Related Work}
\paragraph{Visual Multimodal Generative Models}
The landscape of visual synthesis has been reshaped by the advent of VDMs~\cite{stablevideodiffusion,sora,opensora,opensora2,cogvideo,cogvideox,hunyuanvideo,wan,longcatvideo}, which have established new benchmarks to simulate real-world dynamics.  
Trained on billion-scale datasets, these models possess robust priors beyond the RGB domain. Recent research leverages these priors primarily in two directions: enhancing controllability by incorporating additional visual modalities~\cite{controlnet,t2iadapter,unicontrol,ctrlora,jodi,OmniVDiff,ctrlvdiff,sparsectrl}, and improving perception ability in geometry estimation~\cite{marigold, lotus, depthfm, geowizard, depthcrafter, jointnet, merge, one4d,depthanyvideo,videodepthanything,geometrycrafter} or broader multimodal tasks~\cite{onediffusion,unigeo,geo4d,diception,unityvideo}.
However, this paradigm typically enforces rigid input-output mappings while ignoring the joint correlations shared across modalities. 
Bridging this gap, our work aims to enable versatile video generation by formulating diverse tasks as conditional generation problems within multimodal spaces.

\paragraph{Intrinsic Decomposition and Generation}
Intrinsic image decomposition (inverse rendering), which aims to disentangle RGB images into appearance and geometry-related channels, has long been a fundamental problem in graphics~\cite{bell2014intrinsic}.
Methodologies have evolved from traditional optimization based on physical heuristics~\cite{gkioulekas2013inverse,bonneel2017intrinsic,bousseau2009user,intrinsicscene} to data-driven networks, often tailored for specific domains such as faces~\cite{shu2017neural,shu2018deforming,sun2019single} or complex materials~\cite{wang2022spongecake,li2024tensosdf,physg}. 
Recently, researchers have begun to leverage generative priors to mitigate the ill-posed nature of decomposition~\cite{diffusionrenderer,uni_renderer,intrinsicdiffusion}.
Beyond decomposition, a paradigm of intrinsic generation (text-to-intrinsic) is emerging, shifting to synthesize intrinsic maps directly from text~\cite{lumix,intrinsix,prism}, yet remaining confined to the image level.
In this paper, we introduce \textit{UniVid-Intrinsic} as a representative instantiation of our framework. Unlike prior methods, it enables versatile video generation, where RGB videos and their intrinsic components (albedo, irradiance, normal) can be arbitrarily synthesized from one another or directly from text prompts.
\begin{figure}[t]
    \centering
    \includegraphics[width=\linewidth]{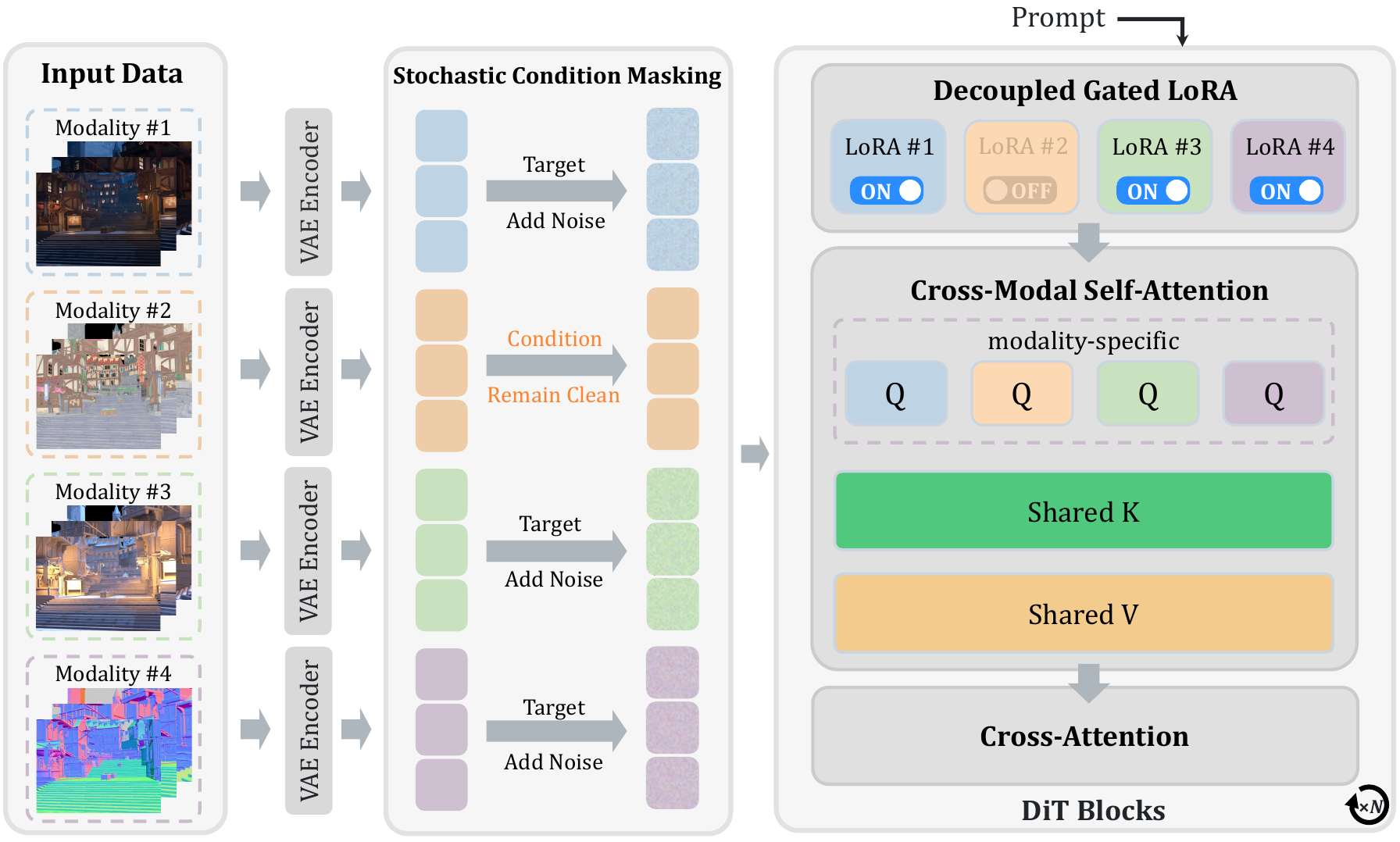}
    \caption{\textbf{Architecture of \textbf{UniVidX} (using \textit{UniVid-Intrinsic} as an example).} Multimodal inputs are encoded and passed through Stochastic Condition Masking (SCM), which randomly assigns them as clean conditions or noisy targets. The DiT blocks are equipped with Decoupled Gated LoRA (DGL): distinct LoRAs are assigned to each modality and are activated only for target inputs while deactivated for conditions (indicated by the faded modules). Modality consistency is ensured via Cross-Modal Self-Attention (CMSA), where queries are modality-specific while keys/values are shared.}
    \label{fig:method}
\end{figure}
\paragraph{Alpha-wise Perception and Generation}
Alpha-channel processing, a cornerstone of computer graphics, has evolved from traditional optimization heuristics~\cite{spectral,closed, tang_learning,designing,chen2007real}, to data-driven paradigms. Modern data-driven approaches have since advanced to precise structure disentanglement, ranging from robust video matting~\cite{chen2018tom,shen2016automatic,rvm,mam,vitmatte,matany,background,bgmv2} to semantic layer decomposition~\cite{semantic,layerdecomp,generative-omnimatte}. 
More recently, a \textit{generative} paradigm has emerged.
Research in this domain has expanded from text-to-RGBA generation~\cite{layerfusion,layerdiffuse,wanalpha} to alpha-guided inpainting, where transparency acts as a spatial constraint for content completion~\cite{propainter,powerpaint,videorepainter}. 
Despite sharing common principles, perception and generation are typically treated in isolation. 
While pioneering efforts like OmniAlpha~\cite{omnialpha} attempt unification at the image level, they rely on specialized alpha-aware VAEs. 
In this paper, we introduce \textit{UniVid-Alpha}. 
By reformulating alpha-wise tasks as conditional video generation, it serves as a representative instantiation of our framework, unlocking versatile capabilities across diverse tasks, including but not limited to video matting, inpainting, and text-to-RGBA generation.
\section{Method}
Our \textbf{UniVidX} is a unified framework designed to leverage the robust VDM priors for versatile multimodal generation. The overall model architecture is illustrated in Fig.~\ref{fig:method}.
In Sec.~\ref{sec:scm}, we introduce \textit{Stochastic Condition Masking (SCM)}, a strategy that breaks the rigidity of fixed input-output mappings by dynamically partitioning modalities into conditions and targets. 
In Sec.~\ref{sec:dgl}, we propose \textit{Decoupled Gated LoRA (DGL)}, which efficiently adapts the backbone to distinct modality distributions without mutual parameter interference. 
In Sec.~\ref{sec:cmsa}, we incorporate \textit{Cross-Modal Self-Attention (CMSA)} to ensure spatiotemporal consistency and dense interaction across diverse modalities. 
Finally, in Sec.~\ref{sec:instantiations}, we detail the implementation of two specific instantiations of \textbf{UniVidX}, namely \textit{UniVid-Intrinsic} and \textit{UniVid-Alpha}, followed by their respective training configurations and dataset strategies in Sec.~\ref{sec:training_data}.
\subsection{Stochastic Condition Masking}
\label{sec:scm}
Video Diffusion Models (VDMs) typically follow a fixed input-output pattern, where the conditional input is restricted to text (T2V) or videos confined to the RGB domain (V2V).
We argue that this rigid distinction between condition and target unnecessarily limits model versatility.
To address this, we propose Stochastic Condition Masking (SCM), a strategy that unifies diverse video tasks into one diffusion model.
Specifically, SCM is built upon a T2V backbone, selected for two strategic reasons: 
(i) it inherently possesses the capability to process pure text inputs, 
and (ii) its latent space is adaptable, allowing us to seamlessly incorporate visual inputs alongside text. 
By dynamically redefining the input-output partition within this fixed multimodal space via SCM, our framework enables versatile video generation for three paradigms: 
Text$\to$X (generating visual modalities from text), 
X$\to$X (translation between visual modalities), 
and Text\&X$\to$X (generation guided by text and visual conditions).

Let $\mathcal{Z}$ denote the collection of latents from all visual modalities. 
During training, we employ a dynamic random partitioning strategy that splits $\mathcal{Z}$ into two mutually exclusive subsets: 1)~Target Subset $\mathcal{Z}_{\text{tgt}}$: The subset selected for generation. These latents serve as the data targets and are corrupted to train the flow model.
2)~Condition Subset $\mathcal{Z}_{\text{cond}}$: The complementary subset. These latents remain clean to serve as conditions for the generation. 
Notably, $\mathcal{Z}_{\text{cond}}$ can be an empty set (e.g., in Text$\to$X tasks, where generation relies solely on text prompts~$c_{\text{txt}}$).

We implement this logical partition via timestep manipulation.
Specifically, for the target subset $\mathcal{Z}_{\text{tgt}}$, we denote the clean latents as $\mathbf{x}^{\mathcal{T}}$. The intermediate noisy state $\mathbf{z}^{\mathcal{T}}_t$ is obtained via linear interpolation between the Gaussian noise $\epsilon \sim \mathcal{N}(0, \mathbf{I})$ and the clean data $\mathbf{x}^{\mathcal{T}}$ at timestep $t \in [0,1]$; 
the latents in $\mathcal{Z}_{\text{cond}}$ are fixed at $t=1$, denoted as $\mathbf{z}_1^{\mathcal{C}}$, serving as unnoised conditions.
Then, the flow matching~\cite{flow_matching} objective $\mathcal{L}_{\text{uni}}$ is formulated to predict the velocity field specifically for the target subset:
\begin{equation}
    \mathcal{L}_{\text{uni}} = \mathbb{E}_{t,\mathbf{x}^{\mathcal{T}},\epsilon} \left\| {\mathbf{v}}_\theta(\mathbf{z}_t^{\mathcal{T}} | \mathbf{z}_1^{\mathcal{C}}, c_{\text{txt}}) - \mathbf{v}\right\|^2_2
    \label{eq:fm_loss}
\end{equation}
where $\theta$ denotes the model parameters. ${\mathbf{v}}_\theta$ is the predicted velocity field, and $\mathbf{v} = \mathbf{x}^{\mathcal{T}} - \epsilon$ corresponds to the ground truth vector field.

This strategy empowers our framework with versatile video generation capabilities. 
During inference, we customize the partition based on specific tasks: latents corresponding to the conditional modalities remain clean to serve as input (or excluded for Text$\to$X), while those for the target modalities are initialized as Gaussian noise. 
This allows for diverse tasks within a single unified model.

\begin{figure*}[t]
    \centering
    \includegraphics[width=\linewidth]{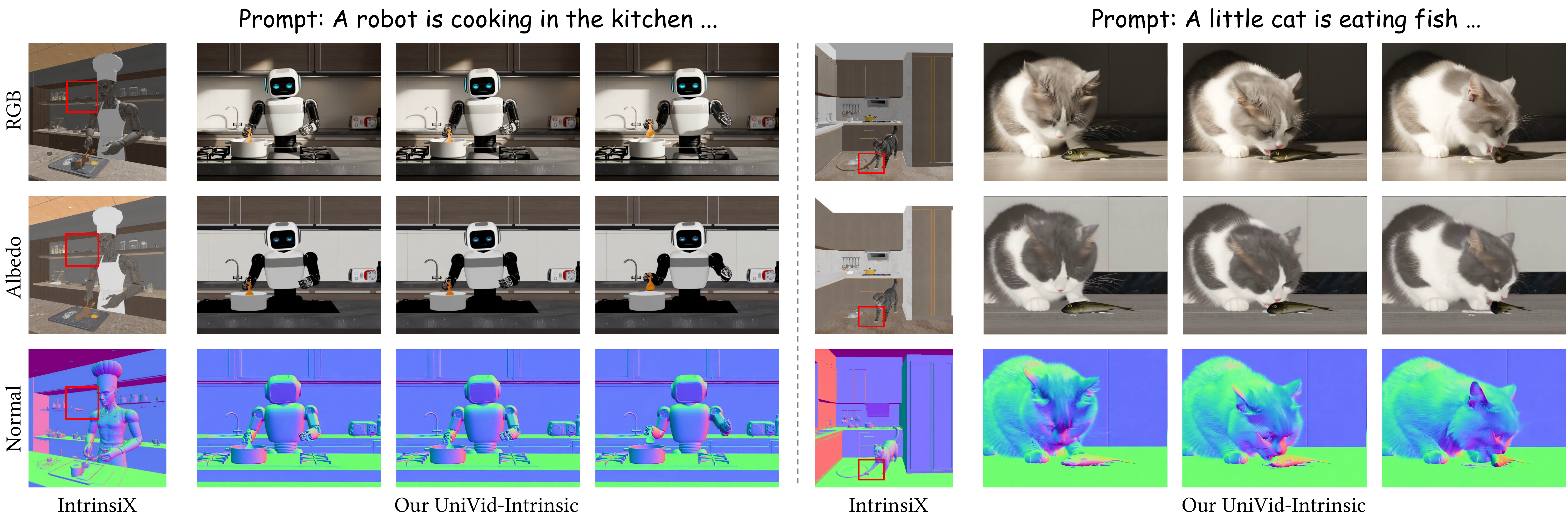}
    \caption{\textbf{Visual comparison for text-to-intrinsic generation.} Compared to IntrinsiX, which exhibits noticeable artifacts and modality misalignment (indicated by \textcolor{red}{red boxes}), our \textit{UniVid-Intrinsic} produces superior results. Our method generates temporally coherent video clips with precise alignment across RGB, albedo, and normal maps, effectively capturing complex geometries and fine textures like the cat's fur. Please zoom in to find more details.}
    \label{fig:text_to_intrinsic}
\end{figure*}

\subsection{Decoupled Gated LoRA}
\label{sec:dgl}
To efficiently leverage the generative priors of pre-trained VDMs while adapting to diverse multimodal requirements, we propose the Decoupled Gated LoRA (DGL) strategy. 
Since different visual modalities follow distinct distributions, sharing parameters across them leads to destructive interference. 
Therefore, instead of applying a monolithic update, DGL assigns independent LoRAs to each specific modality. 
Crucially, these LoRAs are activated only when their corresponding modality serves as a generation target. 
This decoupling effectively prevents parameter interference, allowing the model to capture modality-specific statistics while preserving the robust VDM priors, thereby mitigating the risk of catastrophic forgetting often associated with full fine-tuning, which typically leads to severe performance degradation~\cite{lotus}.

Formally, let $W \in \mathbb{R}^{d \times d}$ denote the frozen pre-trained weights. 
For the $k$-th modality, we introduce a specific parameter update $\Delta W_k = B_k A_k$, where $B_k \in \mathbb{R}^{d \times r}$ and $A_k \in \mathbb{R}^{r \times d}$ are learnable low-rank matrices (r $\mathbin{\ll}$ d). 
This design decouples the processing capabilities for different modalities into distinct parameter spaces, isolating disparate data distributions.
Critically, these LoRAs are dynamically gated based on the role of the modality. 
We formulate the adaptive forward pass to obtain the modality-specific effective weights $W'_k$:
\begin{equation}
    W'_k = W + \mathbf{m}_k \cdot \Delta W_k
\label{eq:DGL}
\end{equation}
When the $k$-th modality serves as a generation target (noisy input), the gate is activated ($m_k=1$); when it serves as a condition (clean input), the gate is suppressed ($m_k=0$), which bypasses the adapter, maximizing the utilization of the VDM's native encoding capability to extract robust semantic features from the visual context without domain-shift interference.
For a detailed analysis of these decoupling and gating designs, please refer to the ablation study in Sec.~\ref{sec:ablation_study}.

\subsection{Cross-Modal Self-Attention}
\label{sec:cmsa}
In our \textbf{UniVidX} framework, data from diverse visual modalities are concatenated along the batch dimension to enable unified processing. However, the vanilla self-attention of standard VDMs operates on each modality in isolation, failing to capture inter-modal dependencies. Motivated by cross-domain diffusion approaches~\cite{intrinsix,wonder3d,wonder3d++,cat3d,viewdiff}, we introduce Cross-Modal Self-Attention (CMSA) to accelerate interaction and fusion across modalities. Specifically, we aggregate the keys and values from all modalities to form a shared context, while keeping the queries modality-specific.

Let $q_i, k_i, v_i$ denote the query, key, and value of the $i$-th modality. 
We construct a shared key/value set by concatenating them: $k_{\text{shared}} = [k_1, k_2,\dots, k_n]$ and $v_{\text{shared}} = [v_1, v_2,\dots, v_n]$. 
The attention operation for modality $i$ is then reformulated as:
\begin{equation}
    \text{Attention}(q_i, k_{\text{shared}}, v_{\text{shared}}) = \text{Softmax}\left(\frac{q_i k_{\text{shared}}^T}{\sqrt{d_k}}\right) v_{\text{shared}}
\label{eq:cm_sa}
\end{equation}

This design ensures that each modality is aware of the multimodal context, thereby promoting cross-modal consistency and enabling alignment between generated content and control conditions.

\subsection{Model Instantiations}
\label{sec:instantiations}
To validate our \textbf{UniVidX}, we implement two instantiations using this framework in two domains. 1) \textit{UniVid-Intrinsic} operates on the RGB videos and their intrinsic maps (albedo/irradiance/normal); 
2) \textit{UniVid-Alpha} focuses on processing blended RGB (BL), alpha mattes (Alpha), foregrounds (FG), and backgrounds (BG). 
Both models operate across three paradigms (Text$\to$X, X$\to$X, and Text\&X$\to$X), supporting a total of \textbf{15} distinct tasks (detailed in the appendix).

\noindent \quad \textit{In \textit{UniVid-Intrinsic} model,} we extend the input space beyond standard RGB videos to capture the underlying physical properties of the scene. 
Specifically, in addition to the RGB video $R \in \mathbb{R}^{T \times H \times W \times 3}$, we incorporate the following intrinsic components: 
1) albedo $A \in \mathbb{R}^{T \times H \times W \times 3}$, representing the surface's diffuse reflectance that remains invariant to illumination and viewing angles; 
2) irradiance $I \in \mathbb{R}^{T \times H \times W \times 3}$, serving as a lighting representation that captures the incoming light intensity accounting for shadows and illumination; and 
3) normal $N \in \mathbb{R}^{T \times H \times W \times 3}$, encoding the per-pixel surface orientation to provide high-frequency geometric details.

While the standard Disney BRDF model~\cite{disneybrdf} characterizes specular reflectance using roughness and metallic maps, we deliberately exclude them from our target modalities. 
This decision is driven by two factors. 
First, reliable ground-truth annotations for material properties are scarce and difficult to curate. Whether synthesized or derived from existing public datasets (e.g., InteriorVerse~\cite{interiorverse}), these labels frequently suffer from significant noise and spatial inconsistency.
Second, we leverage the robust priors of pre-trained VDMs. 
We observe that the VDM possesses an inherent capacity to infer material properties from context, automatically deducing correct material responses to synthesize realistic reflections without needing explicit parameterization.

We also exclude depth maps from our formulation. 
Depth is primarily a macro-geometric attribute rather than a direct photometric component of the shading equation. Moreover, our framework already incorporates surface normals, which capture the finer local geometric details essential for shading computation.

\noindent \quad \textit{In \textit{UniVid-Alpha} model,} we decompose the input video space beyond the blended RGB (BL) video $R \in \mathbb{R}^{T \times H \times W \times 3}$ into three distinct compositing layers: 
1) foreground (FG) $F \in \mathbb{R}^{T \times H \times W \times 3}$, which isolates the intrinsic color and texture details of the subject; 
 2) alpha matte (Alpha) $P \in \mathbb{R}^{T \times H \times W \times 3}$, defining the soft silhouette and per-pixel opacity of the foreground; 
 and 3) background (BG) $B \in \mathbb{R}^{T \times H \times W \times 3}$, capturing the clean environmental context.

The pre-trained VAE encoder in our backbone necessitates 3-channel RGB inputs. 
To ensure compatibility, we adapt the inherently single-channel Alpha by replicating it across three channels before feeding it into the VAE. 
This allows us to process alpha matte within the same latent space as color (RGB).

For the BG layer, we aim to recover the scene as if the foreground subject were never present. 
Leveraging the robust generative capability of the VDM, our model is trained to automatically inpaint regions originally occluded by the foreground. 
This ensures the generation of a spatially complete scene filled with coherent structures and textures, rather than a background with "holes" or artifacts.
\input{tables/text-to-x}
\input{tables/inverse_rendering_and_forward_rendering}
\subsection{Training Details and Data Strategy}
\label{sec:training_data}
\paragraph{Training Details.} 
We build our framework upon the Wan2.1-T2V-14B\footnote{\url{https://huggingface.co/Wan-AI/Wan2.1-T2V-14B}} backbone.  
The rank of LoRA modules in DGL is set to $32$ for all modalities, resulting in a total of $385$M trainable parameters.
We employ a unified optimization strategy for both \textit{UniVid-Intrinsic} and \textit{UniVid-Alpha}, using AdamW~\cite{adamw} ($\beta_1=0.9, \beta_2=0.999$, weight decay=$10^{-2}$) coupled with a Cosine Annealing scheduler~\cite{sgdr} that decays the learning rate from an initial $1 \times 10^{-4}$ to $1 \times 10^{-6}$.

Training is conducted on $4 \times$ NVIDIA H100 GPUs, utilizing BFloat16 (BF16) mixed precision to maximize throughput.
Moreover, both models process video clips of $21$ frames, with a per-GPU batch size of $1$.
Under this setup, \textit{UniVid-Intrinsic} is trained for $6,000$ steps, while \textit{UniVid-Alpha} is trained for $5,000$ steps.

\begin{figure}[t]
    \centering
    \includegraphics[width=\linewidth]{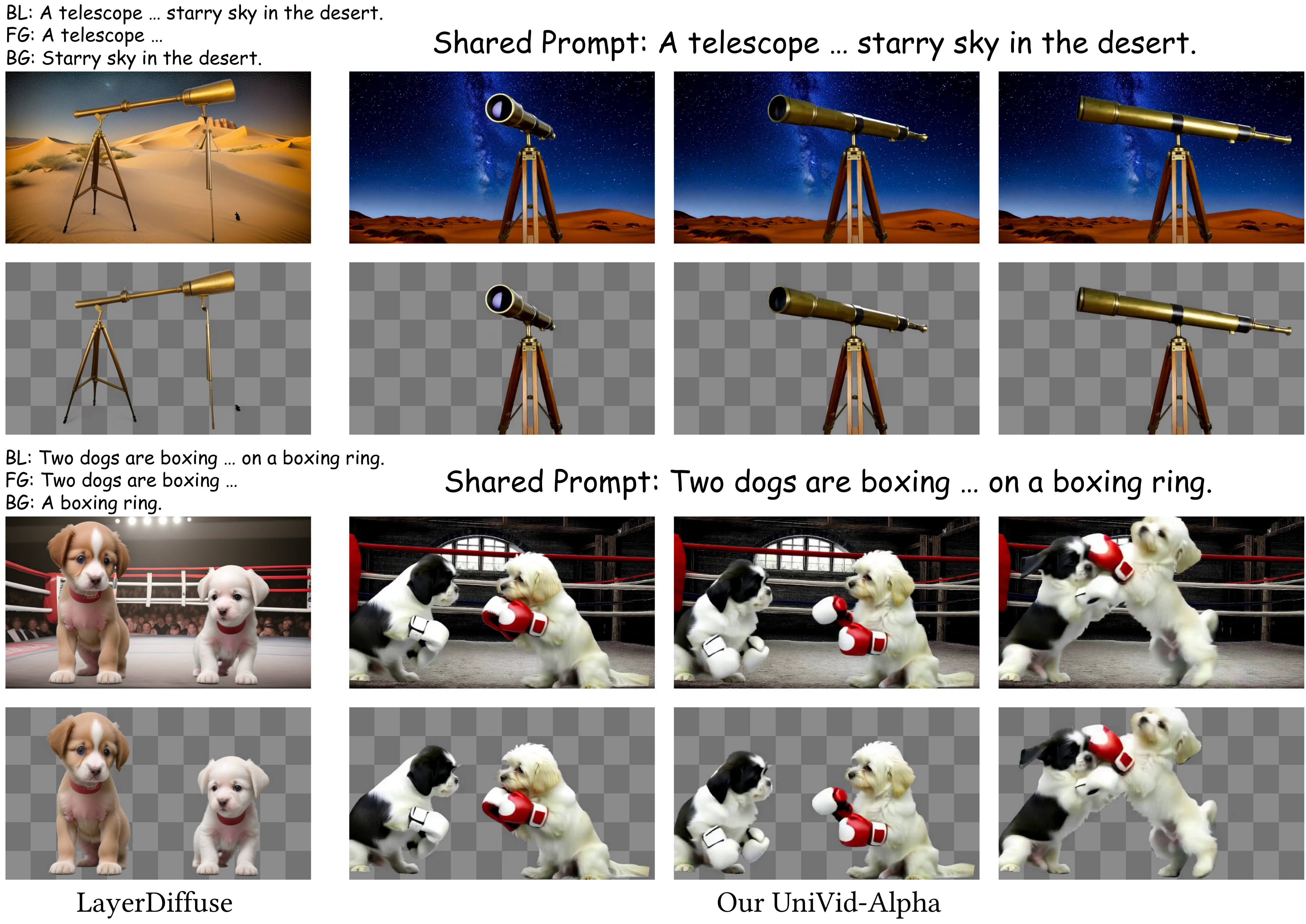}
    \caption{\textbf{Visual results for text-to-RGBA generation.} Compared to LayerDiffuse, which is limited to static images, our method can generate high-quality, dynamic RGBA videos. Notably, while LayerDiffuse needs distinct prompts for different layers to ensure separation, our method achieves robust performance using a single shared prompt.}
    \label{fig:text_to_rgba}
\end{figure}

\paragraph{Training Dataset.}
For \textit{UniVid-Intrinsic}, we require high-quality RGB videos paired with ground-truth albedo, irradiance, and normal maps.
Since such dense physical supervision is unattainable in real-world data and existing public synthetic datasets typically provide only a subset of these modalities, we construct a synthetic dataset \textsc{InteriorVid}.
It comprises $924$ high-quality indoor video clips, each consisting of $21$ frames at a resolution of $480 \times 640$, with paired ground-truth for albedo, irradiance, and normal maps (see appendix for construction details).
We partition the dataset into \textsc{InteriorVid-Train} ($900$ clips) for training and \textsc{InteriorVid-Test} ($24$ clips) for testing.
For \textit{UniVid-Alpha}, we utilize VideoMatte240K~\cite{bgmv2}, a widely adopted dataset for video matting featuring human foregrounds with paired ground-truth alpha mattes.
We use 484 videos from this dataset to train our model, with resolution resized to $432 \times 768$.
To obtain text descriptions, we leverage Qwen3-VL~\cite{Qwen3-VL} to generate captions for the training data.

\paragraph{Construction Details of \textsc{InteriorVid}.}
To construct \textsc{InteriorVid}, we curate $167$ high-quality 3D indoor scenes from SuperHiveMarket\footnote{\url{https://superhivemarket.com/}}.
To simulate realistic camera dynamics, we implement smooth random walk trajectories for each scene, further augmented with randomized Field of View (FOV) and focal lengths. 
This setup ensures that the resulting dataset encompasses a diverse array of motion patterns and perspective variations.

The data generation pipeline is executed using Blender\footnote{\url{https://www.blender.org/}} with the Cycles path-tracing engine ($128$ samples).We implement a fine-grained decoupling of physical components via the Blender Compositor node tree.
Crucially, all output components are exported in OpenEXR 16-bit Float format to preserve the full dynamic range in linear space, strictly ensuring that the decomposed layers adhere to the constraints of the physical rendering equation.

\section{Experiment}
In this section, we provide a detailed experimental analysis of our framework.
We first outline the experimental setup, detailing the specific tasks evaluated for both models (Sec.~\ref{sec:exp_setup}). 
Next, we provide comprehensive qualitative and quantitative comparisons against other baselines (Sec.~\ref{sec:com_eva}). 
Specifically, we detail the results for text-to-intrinsic and text-to-RGBA generation in Sec.~\ref{sec:text2x}. Evaluations for inverse/forward rendering are presented in Sec.~\ref{sec:inverse_forward_rendering}. 
We further report albedo estimation results in Sec.~\ref{sec:albedo_estimation}, and we also include a focused assessment of normal estimation in Sec.~\ref{sec:normal_estimation}.  
Finally, we demonstrate our video matting performance in Sec.~\ref{sec:video_matting}. 

We then conduct thorough ablation studies to validate the effectiveness of our core architectural designs (Sec.~\ref{sec:ablation_study}).
In Sec.~\ref{sec:discussion}, we discuss the critical value of multi-condition perception in resolving ambiguity. 
Furthermore, we demonstrate the flexibility of our framework, illustrating how the composition of different tasks supports diverse downstream applications (Sec.~\ref{sec:application}). 
Finally, we analyze the current limitations and failure cases in Sec.~\ref{sec:limitations}.

\subsection{Experimental Setup}
\label{sec:exp_setup}
We focus on representative tasks that allow for quantitative comparison. 
For \textit{UniVid-Intrinsic}, we evaluate: 
(1) text-to-intrinsic (Text$\to$X), which jointly generates RGB videos and their corresponding intrinsic maps from text prompts; 
(2) inverse rendering (X$\to$X), which estimates intrinsic maps given an input RGB video, including dedicated evaluations of albedo/normal estimation as critical sub-tasks; 
and (3) forward rendering (X$\to$X), which performs realistic RGB video synthesis derived from input intrinsic channels.
For \textit{UniVid-Alpha}, we evaluate: 
(1) text-to-RGBA (Text$\to$X), which synthesizes decomposed RGBA layers and the final blended video from text; 
and (2) video matting (X$\to$X), which decomposes an input blended video into its constituent RGBA layers. 

\begin{figure}
    \centering

    \begin{subfigure}{\linewidth}
        \centering
        \includegraphics[width=\linewidth]{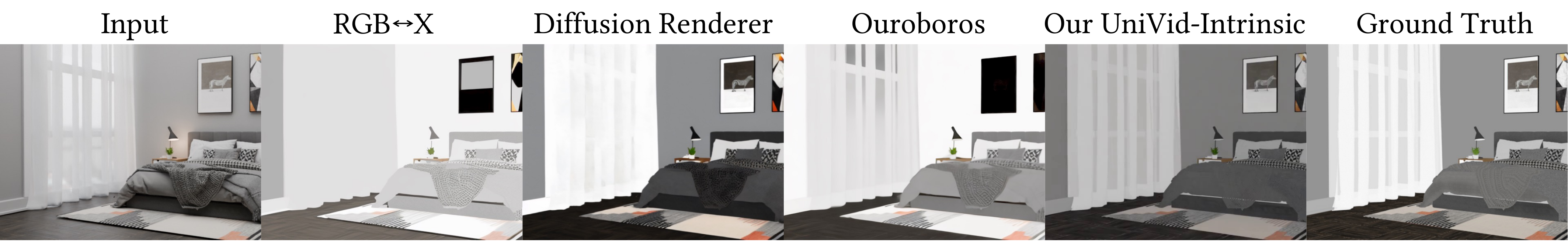}
        \caption*{\scriptsize (a)~Albedo estimation. Comparison of estimated albedo maps.}
        \label{fig:albedo}
    \end{subfigure}
    \begin{subfigure}{\linewidth}
        \centering
        \includegraphics[width=\linewidth]{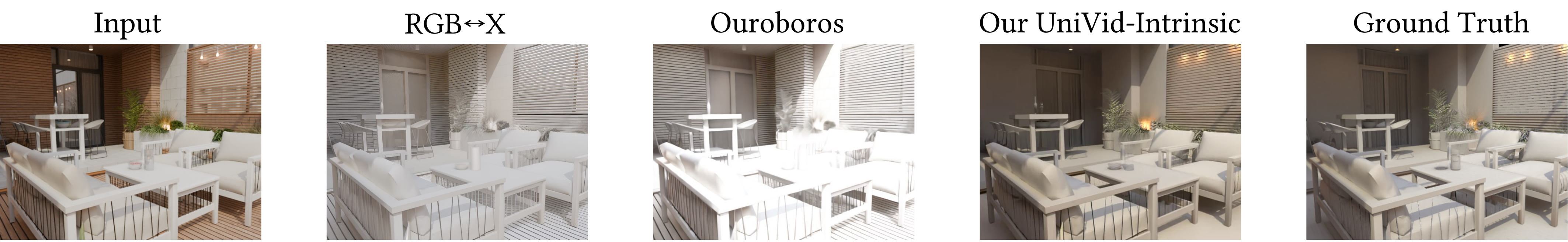}
        \caption*{\scriptsize (b)~Irradiance estimation. Comparison of estimated irradiance maps.}
        \label{fig:irradiance}
    \end{subfigure}
    \begin{subfigure}{\linewidth}
        \centering
        \includegraphics[width=\linewidth]{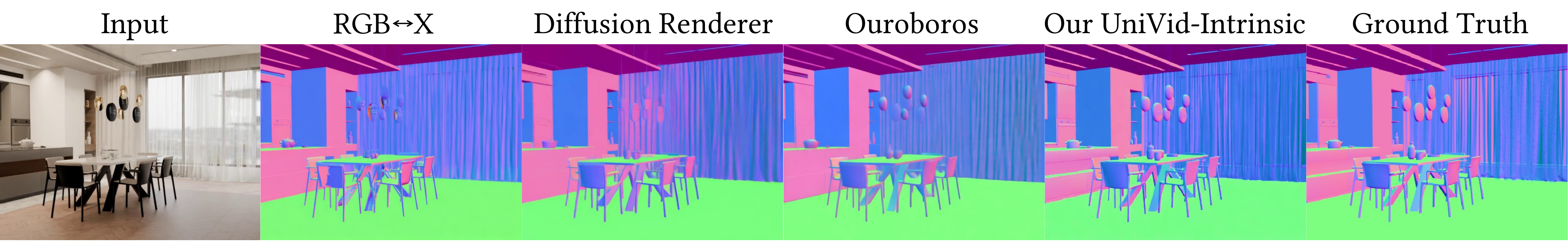}
        \caption*{\scriptsize (c)~Normal estimation. Comparison of estimated normal maps.}
        \label{fig:normal_}
    \end{subfigure}
    \begin{subfigure}{\linewidth}
        \centering
        \includegraphics[width=\linewidth]{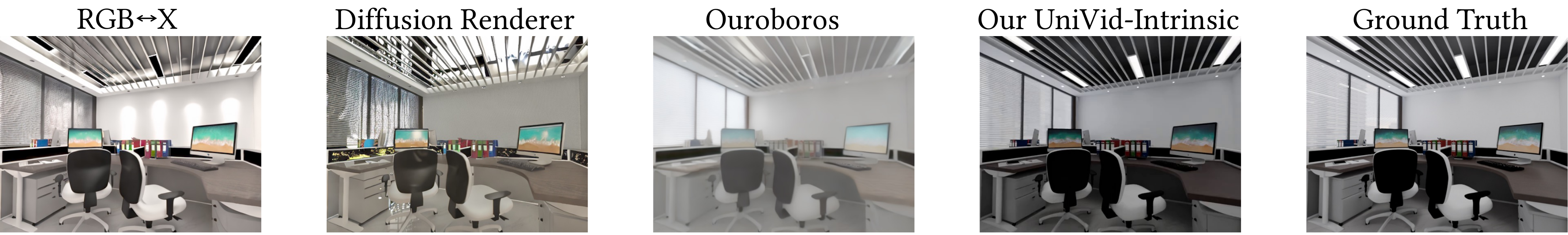}
        \caption*{\scriptsize (d)~Forward rendering. Comparison of reconstructed RGB videos.}
        \label{fig:forward_rendering}
    \end{subfigure}
    \caption{\textbf{Visual comparison for inverse and forward rendering tasks.} In all tasks, \textit{UniVid-Intrinsic} produces results closest to the Ground Truth.}
    \label{fig:inverse_rendering_and_forward_rendering}
\end{figure}

\begin{figure*}[t]
    \centering
    \includegraphics[width=0.97\linewidth]{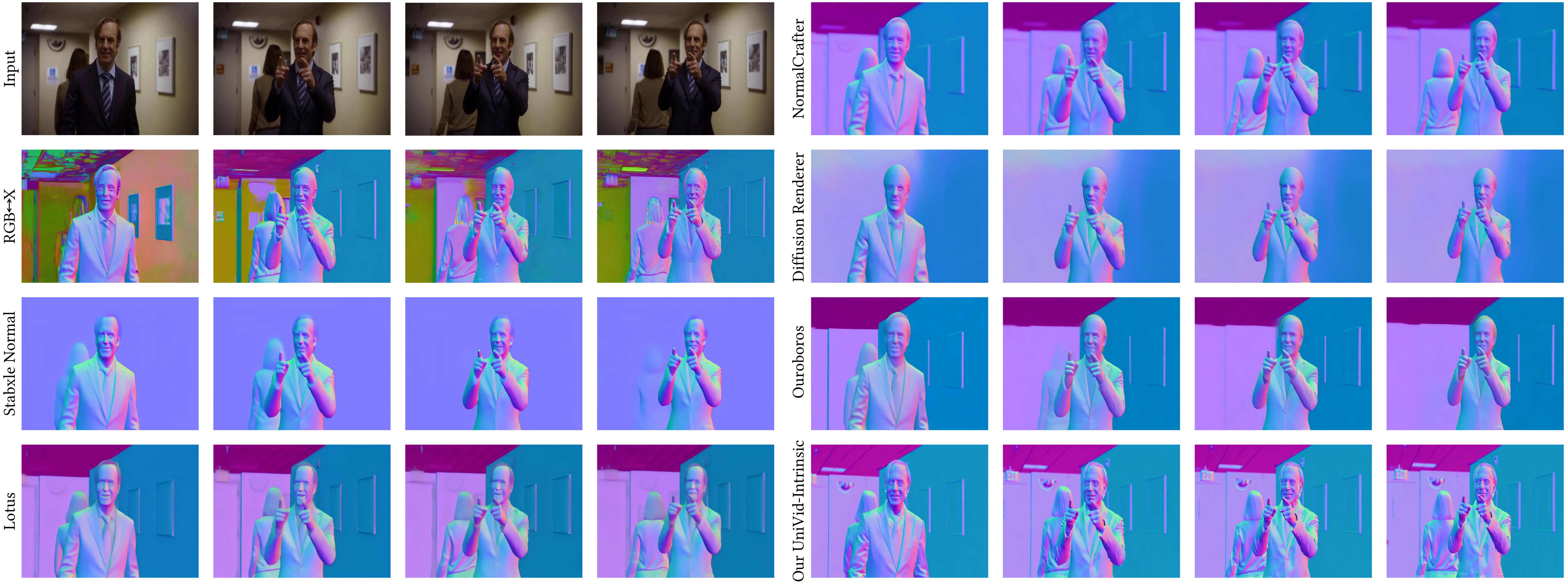}
    \caption{\textbf{Normal estimation on a cinematic video sequence.} Compared to specialized normal estimators and intrinsic-related baselines which struggle with temporal stability or detail preservation, our method yields temporally coherent normals while maintaining high-fidelity geometric details.}
    \label{fig:normal}
\end{figure*}

\subsection{Comparative Evaluation}
\label{sec:com_eva}
\subsubsection{Text$\to$X}
\label{sec:text2x}
\input{tables/albedo_estimation}
Due to the absence of open-source text-to-video methods for text-to-intrinsic and text-to-RGBA, we benchmark our methods against representative image generation models.
For text-to-intrinsic, we compare \textit{UniVid-Intrinsic} against IntrinsiX~\cite{intrinsix} on the intersection of modalities: RGB, albedo, and normal. 
Notably, while our model generates RGB frames simultaneously with intrinsic maps, the RGB images of IntrinsiX are rendered from its generated intrinsic maps following its official protocol.
For text-to-RGBA, we compare \textit{UniVid-Alpha} against LayerDiffuse~[Zhang et al.~\citeyear{layerdiffuse}]. 
Both methods take text as input to generate foreground (FG), background (BG), and the blended RGB (BL) result.

To assess generation quality, we conduct a user study where participants rate results on a scale from 1 to 10. 
We utilized Gemini 3 Pro~\footnote{\url{https://gemini.google.com/}} to design evaluation prompts, resulting in $221$ samples for both tasks. 
Evaluation criteria include (1) visual quality (of all generated modalities), (2) text alignment (TA), and (3) modality consistency (MC).
Furthermore, given the temporal nature of our outputs, we employ the Temporal Flickering metric (range 0-1, higher is better) from VBench~\cite{vbench} to evaluate temporal stability.

Across both text-to-intrinsic and text-to-RGBA tasks, our \textit{UniVid-Intrinsic} and \textit{UniVid-Alpha} consistently surpass the representative baselines (IntrinsiX and LayerDiffuse, respectively). 
In user studies~(see Tab.~\ref{tab:combined_results}), we obtain higher ratings for visual quality, text alignment, and modality consistency.
Furthermore, our Temporal Flickering scores are consistently close to $1.0$, confirming our ability to generate temporally stable content, which is a critical advantage over image-based baselines.

\input{tables/appendix_sintel}
The qualitative results validate the effectiveness of our method. 
1) For text-to-intrinsic, while IntrinsiX often exhibits misalignment among modalities (highlighted by red boxes in Fig.~\ref{fig:text_to_intrinsic}), our \textit{UniVid-Intrinsic} maintains consistency. 
Additionally, we excel in generating realistic illumination (see Fig.~\ref{fig:text_to_intrinsic} row 1) and high-frequency geometry details, such as the fur of the cat (see Fig.~\ref{fig:text_to_intrinsic} row 3). 
2) For text-to-RGBA, despite being trained only on dataset~\cite{bgmv2} significantly smaller than LayerDiffuse ($484$ videos vs $1$M images), and without requiring VAE fine-tuning, the generation quality of our \textit{UniVid-Alpha} remains impressive, demonstrating the effectiveness of leveraging VDM priors. 
Furthermore, unlike LayerDiffuse, which relies on distinct prompts for the BL, FG, and BG layers to ensure quality, our method achieves robust performance using a shared prompt. 
This is attributed to decoupling design in DGL~(please refer to Sec.~\ref{sec:ablation_study} for details).
Moreover, although both models are trained on limited domain-specific data~(\textit{UniVid-Intrinsic} on indoor scenes; \textit{UniVid-Alpha} on human data), they generalize well to out-of-distribution samples, such as animals.

\subsubsection{Inverse Rendering and Forward Rendering}
\label{sec:inverse_forward_rendering}
We benchmark \textit{UniVid-Intrinsic} on inverse and forward rendering tasks against several representative methods like RGB$\leftrightarrow$X~\cite{rgbx}, Diffusion Renderer~\cite{diffusionrenderer} and Ouroboros~\cite{ouroboros}. 
For normal estimation, we include comparisons with specialized normal estimation methods: Stable Normal~\cite{stablenormal}, Lotus~\cite{lotus}, and NormalCrafter~\cite{normalcrafter}. 
All evaluations are conducted on the \textsc{InteriorVid-Test} benchmark (see Sec.~\ref{sec:training_data}).

To quantitatively evaluate performance, we measure PSNR, SSIM, and LPIPS on both the estimated intrinsic maps (inverse rendering) and the reconstructed RGB videos (forward rendering).
For surface normals, we report geometric accuracy using the Mean Angular Error (MAE) and the percentage of pixels with errors below $11.25^\circ$.

Both quantitative and qualitative results demonstrate that our \textit{UniVid-Intrinsic} achieves state-of-the-art performance. 
Quantitatively (see Tab.~\ref{tab:inverse_rendering_and_forward_rendering}), our method not only outperforms intrinsic baselines, but also surpasses specialized estimators (e.g., Stable Normal) in surface normal estimation, achieving the lowest MAE of $11.09^\circ$. 
Qualitatively (see Fig.~\ref{fig:inverse_rendering_and_forward_rendering}), our method produces results that most closely resemble the ground truth. Specifically, it recovers artifact-free albedo (row 1), illumination-consistent irradiance maps (row 2), and high-quality normal maps (row 3) in inverse rendering, alongside high-fidelity reconstruction (row 4) in forward rendering.

\subsubsection{Albedo Estimation}
\label{sec:albedo_estimation}
Albedo estimation has long been a fundamental problem in graphics. 
To further evaluate the performance of our method, particularly its transfer to real-world scenes, we report results on the Measured Albedo in the Wild (MAW) dataset~\cite{wu2023measured}. 
MAW is a real-world benchmark for albedo estimation that measures accuracy in terms of both intensity and chromaticity. 
It consists of ~$850$ images, each annotated with measured albedo in specific masked regions, where the measurements are obtained using a known gray card placed on areas of homogeneous albedo.

As shown in Tab.~\ref{tab:albedo_estimation}, our \textit{UniVid-Intrinsic} achieves the best intensity error of $0.44$ and a competitive chromaticity error of $3.60$, placing it among the top-performing methods. 
Notably, although \textit{UniVid-Intrinsic} is trained solely on synthetic data, it transfers well to this real-world benchmark, suggesting promising generalization ability.

\input{tables/video_matting}
\subsubsection{Normal Estimation}
\label{sec:normal_estimation}
Given the critical role of geometry in scene understanding, we provide a focused analysis of our normal estimation capabilities.
As shown in Fig.~\ref{fig:normal}, while these baselines frequently suffer from texture loss and temporal flickering, our method faithfully recovers high-frequency details (e.g., facial features) and ensures temporally consistent results free from jitter.

Quantitatively, we present an evaluation of \textit{UniVid-Intrinsic} against state-of-the-art specialized normal estimation models on the Sintel~\cite{sintel} benchmark. 
Our comparison set encompasses robust image-based methods, including DSINE~\cite{dsine}, GeoWizard~\cite{geowizard}, GenPercept~\cite{genpercept}, Stable-Normal~\cite{stablenormal}, Marigold-E2E-FT~\cite{e2eft}, and Lotus~\cite{lotus}, as well as the video-based baseline NormalCrafter~\cite{normalcrafter}. Following standard evaluation protocols, we report the Mean and Median angular errors ($\downarrow$), alongside the accuracy within angular thresholds of $11.25^{\circ}$, $22.5^{\circ}$, and $30^{\circ}$ ($\uparrow$).
Additionally, we explicitly report the training data scale for each method to analyze data efficiency.

\begin{figure}[t]
    \centering
    \includegraphics[width=\linewidth]{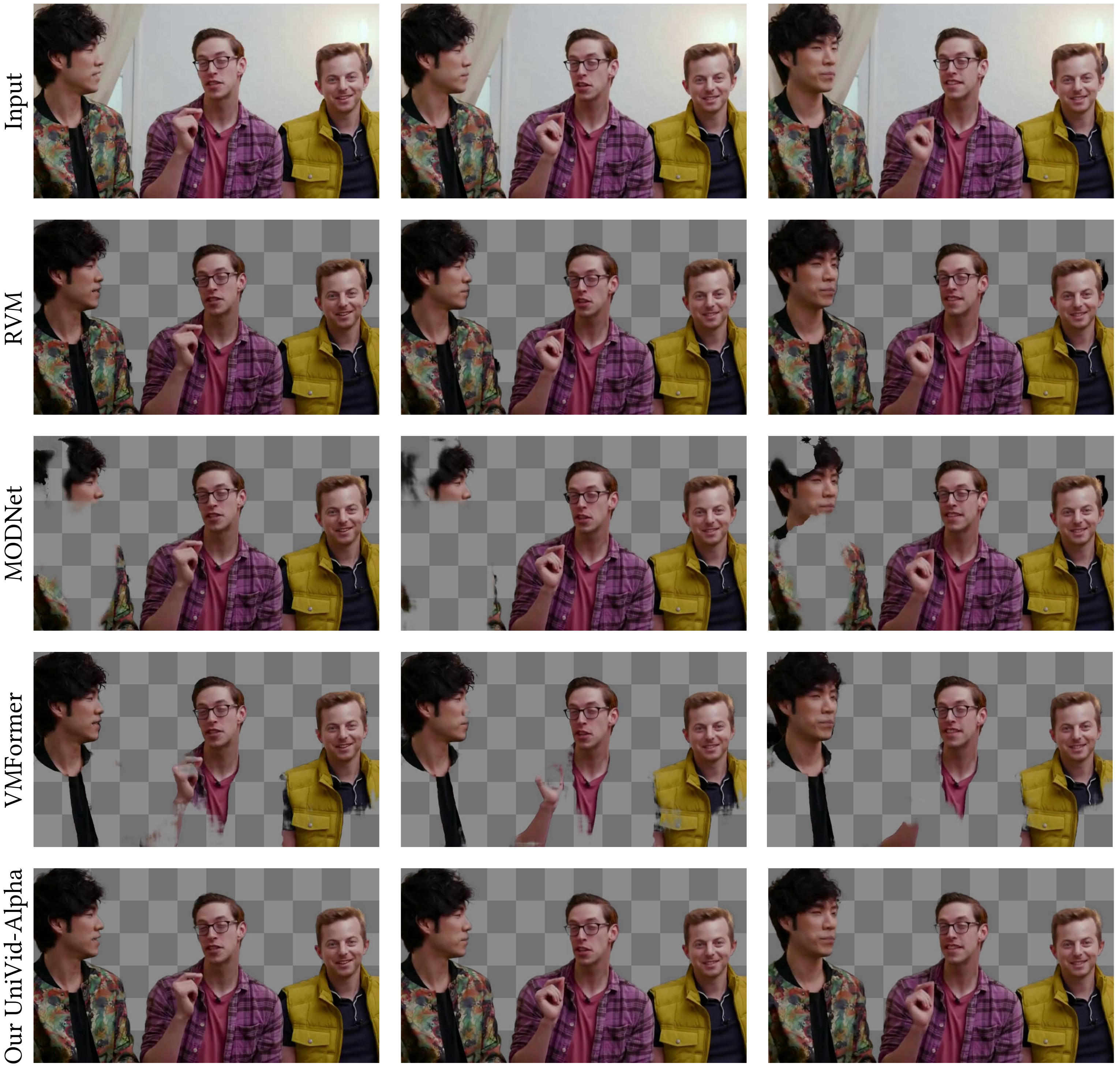}
    \caption{\textbf{Visual comparison of auxiliary-free video matting results.} While competing approaches exhibit noticeable artifacts and background leakage (e.g., the wall sconce), our method produces accurate mattes.}
    \label{fig:video_matte}
\end{figure}

As shown in Tab.~\ref{tab:sintel}, \textit{UniVid-Intrinsic} achieves performance comparable to these specialized baselines while requiring significantly less training data. 
Notably, compared to the video-specific counterpart NormalCrafter~\cite{normalcrafter}, our model demonstrates superior data efficiency: we utilize only 19K training frames compared to their 860K (\textbf{a reduction of over 45$\times$}). 
This highlights that our framework effectively leverages strong diffusion priors, enabling robust generalization even when trained on small-scale datasets.
\begin{figure*}[t]
    \centering
    \includegraphics[width=\linewidth]{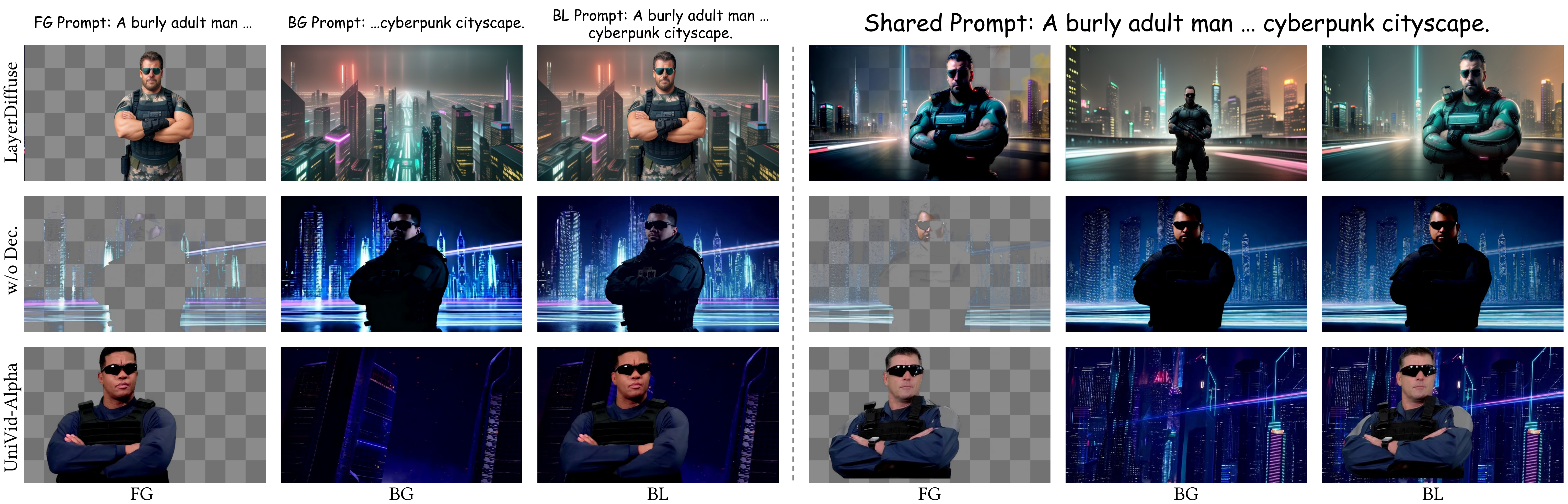}
    \caption{\textbf{Qualitative ablation of decoupling design.} Comparison of generation using distinct prompts (Left) vs. a shared prompt (Right). While LayerDiffuse fails with shared prompt and the 'w/o Dec.' variant consistently fails due to parameter sharing, our approach achieves robust generation in both situations.}
    \label{fig:wo_decoupled}
\end{figure*}
\subsubsection{Video Matting}
\label{sec:video_matting}
For video matting, our \textit{UniVid-Alpha} operates as an Auxiliary-Free (AF) method, requiring only RGB inputs.
We compare our approach against two categories of video matting methods: AF Methods, such as RVM~\cite{rvm}, MODNet~\cite{MODNet}, and VMFormer~\cite{vmformer}; Mask-Guided (MG) Methods, such as AdaM~\cite{lin2023adam}, FTP-VM~\cite{FTP_VM}, MaGGIe~\cite{maggie} and MatAnyone~\cite{matanyone}, which require additional segmentation masks as inputs.

For quantitative evaluation, we employ MAD (Mean Absolute Difference) and MSE (Mean Squared Error) to assess semantic accuracy, Grad~\cite{2009perceptually} for detail extraction, dtSSD~\cite{2015perceptually} for temporal coherence, and Conn (Connectivity)~\cite{2009perceptually} for perceptual quality. Quantitative evaluations are conducted on the VideoMatte~\cite{bgmv2} benchmark.

Quantitatively, while MG methods typically outperform AF methods due to explicit guidance, our method defies this trend. 
As shown in Tab.~\ref{tab:youtubematte}, we achieve state-of-the-art results (e.g., lowest MAD of 4.24), outperforming both AF and MG competitors. Qualitatively (Fig.~\ref{fig:video_matte}), this advantage is evident in challenging multi-subject in-the-wild scenarios. 
Although competing approaches suffer from significant artifacts and background leakage (e.g., the wall sconce), our method produces clean, coherent mattes, accurately preserving even intricate hair details. This success stems from our effective use of VDM priors, which provide the robust semantic segmentation capability needed to distinguish subjects from complex backgrounds without auxiliary inputs~\cite{segdiff,diffseg}.

It is worth highlighting that while traditional video matting methods are typically limited to yielding only foregrounds and alpha mattes, our \textit{UniVid-Alpha} leverages the generative capabilities of VDMs to jointly synthesize a clean background.
\begin{figure}[t]
    \centering
    \includegraphics[width=\linewidth]{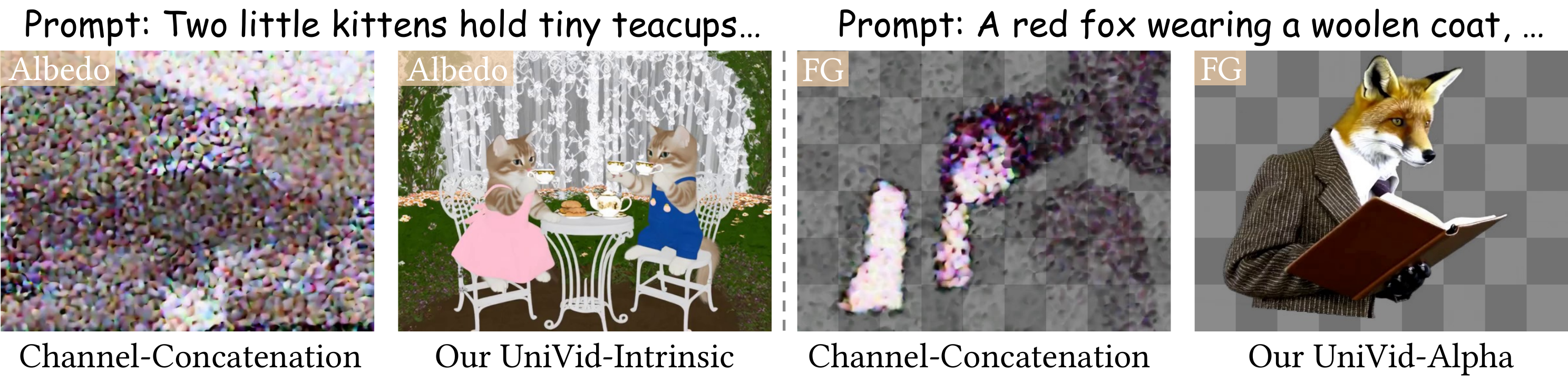}
    \caption{\textbf{Visual results of channel-concatenation.} Left: Albedo results for text-to-intrinsic generation. Right: FG results for text-to-RGBA generation. In both tasks, the channel-concatenation variant fails completely, yielding corrupted outputs due to the disruption of the diffusion priors. Conversely, \textit{UniVid-Intrinsic} and \textit{UniVid-Alpha} models generate high-fidelity results, demonstrating the superiority of our \textbf{UniVidX}.
    }
    \label{fig:w_channel_concat}
\end{figure}

\begin{figure}[t]
    \centering
    \includegraphics[width=\linewidth]{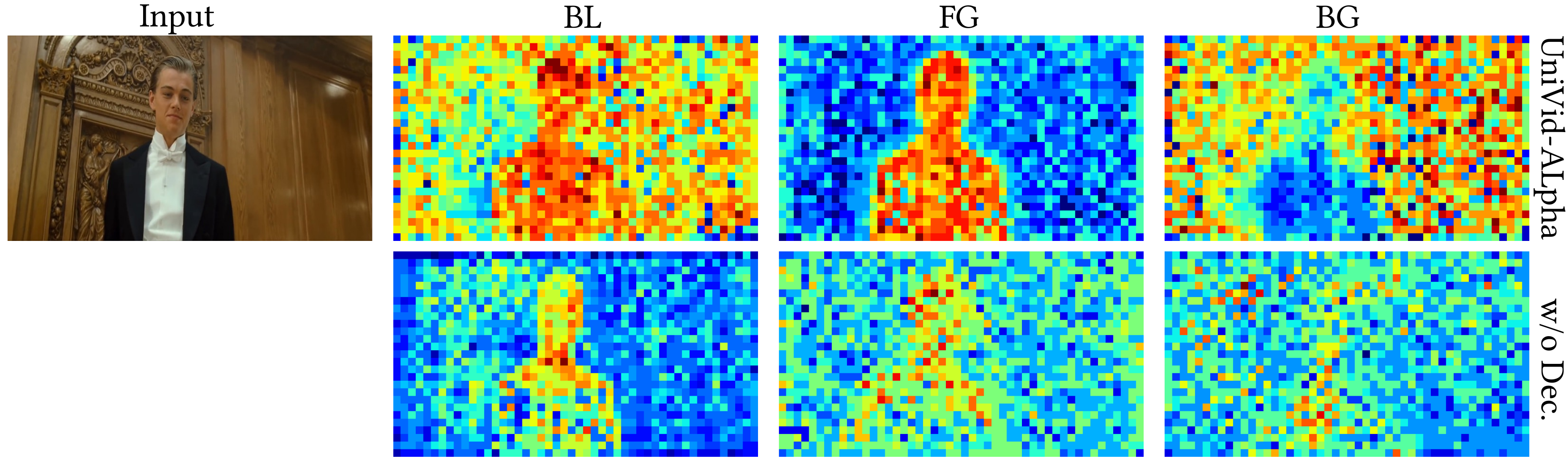}
    \caption{\textbf{Attention map analysis.} Maps are extracted from the Cross-Modal Self-Attention layers in the 20th DiT block at denoising step 25/50. Top: Our method yields clean attention maps where FG and BG branches distinctively attend to the subject and background. Bottom: The 'w/o Dec.' variant results are noisy, proving its inability to separate different modalities effectively.}
    \label{fig:attn_map}
\end{figure}
\begin{figure}[t]
    \centering
    \includegraphics[width=\linewidth]{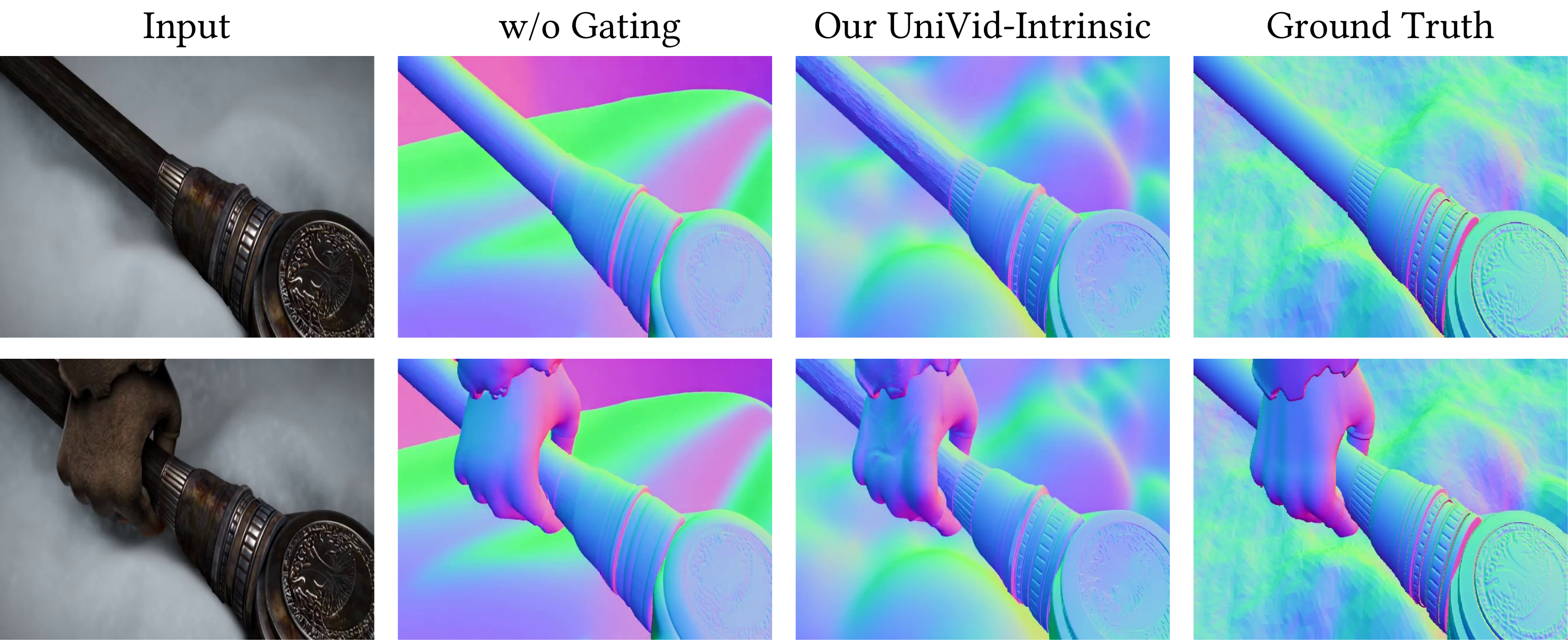}
    \caption{\textbf{Qualitative ablation of the gating design.} While the 'w/o Gating' variant suffers from inaccurate background prediction and texture loss, our full model demonstrates robust normal estimation capabilities.}
    \label{fig:wo_gated}
\end{figure}

\subsection{Ablation Study}
\label{sec:ablation_study}

\begin{figure*}[t]
    \centering
    \includegraphics[width=0.97\linewidth]{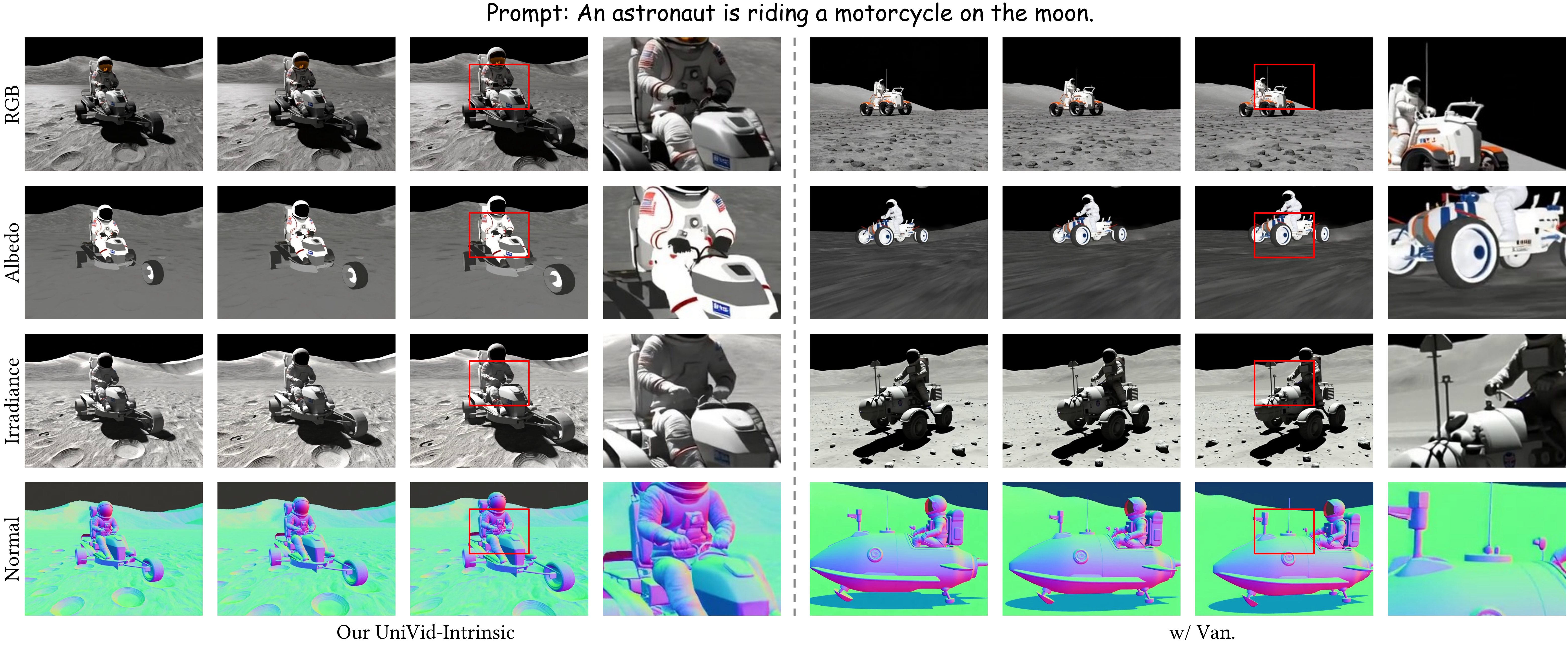}
    \caption{\textbf{Qualitative ablation on Cross-Modal Self-Attention.} We compare the text-to-intrinsic generation results of our \textit{UniVid-Intrinsic} with the 'w/ Van.' variant using the same prompt. As shown, our model demonstrates superior structural consistency across all modalities (RGB, albedo, irradiance, and normal). In contrast, the 'w/ Van.' variant suffers from noticeable inconsistencies and misalignment between different modalities.}
    \label{fig:wo_attn}
\end{figure*}

\noindent \quad \textit{Why do we not use channel-concatenation?} To enable simultaneous multimodal generation, a prevalent paradigm is channel-concatenation, adopted by methods like Diffusion Render~\cite{diffusionrenderer}, Geo4D \cite{geo4d}, and CtrlVDiff~\cite{ctrlvdiff}.
This approach stacks latents from different modalities along the channel dimension before feeding them into the DiT.
While theoretically advantageous for preserving spatial correspondence and pixel alignment, we find that this strategy severely compromises the pre-trained diffusion priors.
The necessity of retraining input convolutional layers from scratch and adding new output heads causes a significant shift in the internal feature distribution.
Although previous works mitigate this by training on massive datasets (e.g., $\sim$350K videos in CtrlVDiff), our experiments reveal that this method fails under limited data regimes.
To verify this, we trained variants of both \textit{UniVid-Intrinsic} and \textit{UniVid-Alpha} using the channel-concatenation strategy.
As shown in Fig.~\ref{fig:w_channel_concat}, the generated videos from these variants suffer from severe structural collapse.
In contrast, our \textbf{UniVidX} concatenates multimodal latents along the batch dimension.
This approach requires no modifications to the input/output structures, thereby maximally leveraging native VDM priors and achieving superior data efficiency ($<1$K videos).
Consequently, as evident in Fig.~\ref{fig:w_channel_concat}, our model produces high-fidelity results, effectively overcoming the collapse observed in the channel-concatenation variants.

\noindent \quad \textit{Why do we need decoupling design in DGL?}~In our Decoupled Gated LoRA strategy, we assign an independent LoRA module to each specific modality. 
This design is intended to decouple the processing capabilities for distinct modalities into separate parameter spaces, thereby significantly enhancing training robustness.
\input{tables/ablation_wo_gated}
To validate the necessity of this decoupling strategy, we conduct an ablation study on \textit{UniVid-Alpha} by comparing our method against a shared-parameter variant.
For a fair comparison, we implement a shared LoRA variant (named 'w/o Dec.') instead of full fine-tuning and set the rank of the shared LoRA to $64$ (double that of our decoupled modules) to maintain an identical parameter count. 
Furthermore, we add distinct RoPE~\cite{rope} positional encoding to different modalities in the 'w/o Dec.' setup. 
Conversely, since the decoupling mechanism inherently handles modality distinction, our model utilizes identical positional encoding for all modalities.

As shown in Fig.~\ref{fig:attn_map}, our method exhibits clear modality disentanglement: the BL branch focuses globally, the FG branch concentrates precisely on the foreground subject, and the BG branch covers the background. 
In contrast, the 'w/o Dec.' variant produces chaotic and noisy attention maps, exhibiting severe feature leakage across FG and BG. This indicates that without parameter decoupling, the model struggles to effectively differentiate between modalities.

In text-to-RGBA task, our method maintains robust layer separation with both specific and shared prompts. 
Conversely, the 'w/o Dec.' variant suffers from severe foreground-background confusion in both scenarios. 
Notably, we observe that LayerDiffuse~[Zhang et al.~\citeyear{layerdiffuse}], which also relies on shared parameters, fails to separate layers when using a shared prompt.
This comparison reinforces that the decoupling design is critical for robust multimodal processing.

\textit{Why do we need gating design in DGL?}
To prevent task-specific parameters from interfering with the backbone's native encoding capabilities, we employ a gating mechanism. 
This strategy selectively activates LoRAs only when a modality serves as generation target (noisy input) and deactivates them when it serves as condition (clean input). 
We validate this design by comparing our \textit{UniVid-Intrinsic} against a "w/o Gating" model, where the gating logic is disabled by fixing $\mathbf{m}_k = 1$ in Eq.~\ref{eq:DGL} to keep LoRA permanently active. Qualitatively (see Fig.~\ref{fig:wo_gated}), the 'w/o Gating' variant suffers from low-quality snowy ground normal estimation and severe texture loss on the walking stick.
This is further corroborated by quantitative evaluations on \textsc{InteriorVid-Test} (Tab.~\ref{tab:ablation_gated_lora}), where the variant underperforms \textit{UniVid-Intrinsic}. For example, the albedo PSNR drops to $15.02$ dB, a decrease of $1.87$ dB.
Collectively, these results confirm that the gated mechanism is essential for utilizing VDM's priors.

\begin{figure}[t]
    \centering
    \includegraphics[width=\linewidth]{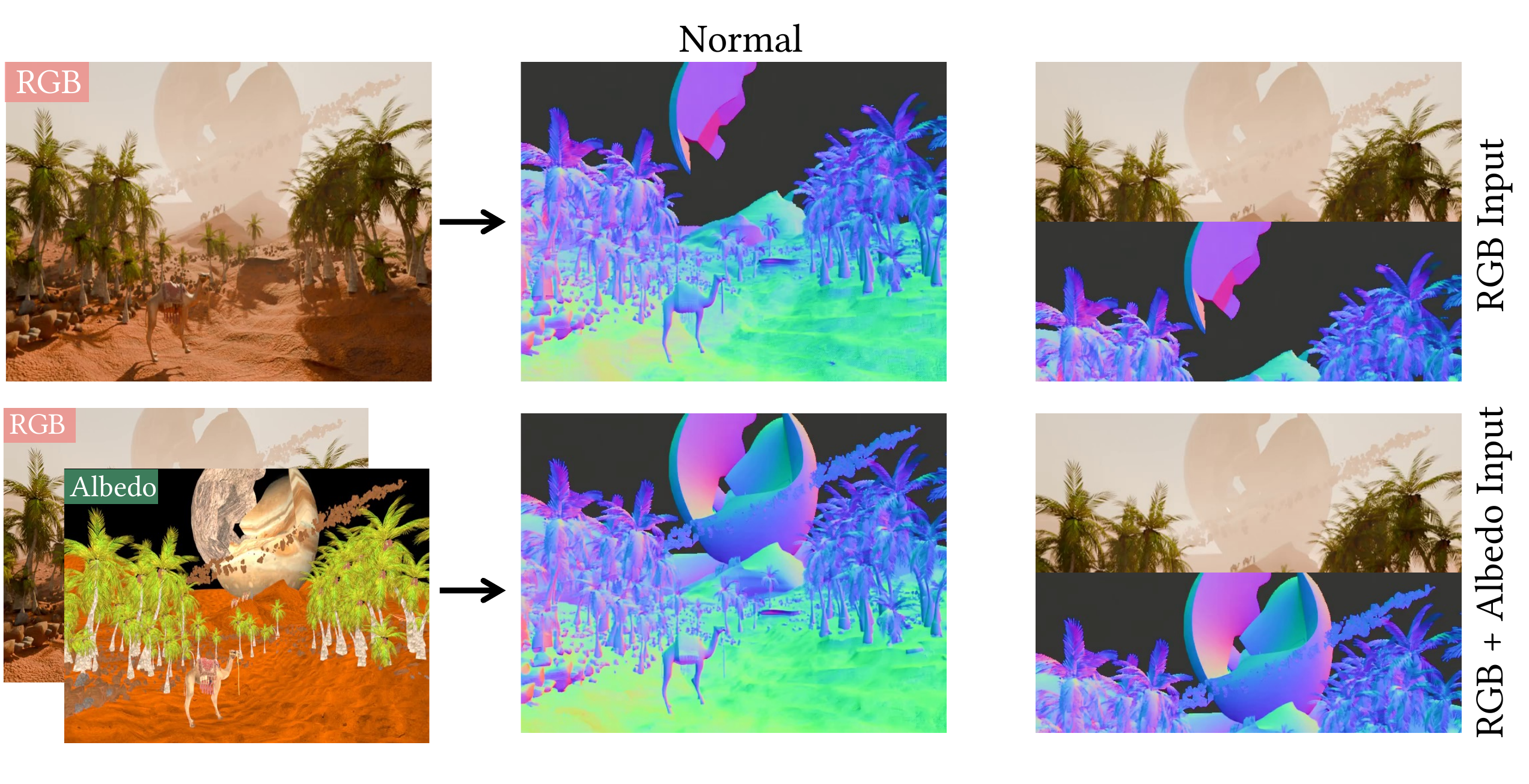}
    \caption{\textbf{Demonstrating the value of multi-condition for mitigating perceptual ambiguity.} The single-condition RGB input (top) fails to capture the geometry of the distant, blurry object due to the inherent ambiguity of the RGB input. In contrast, by utilizing the auxiliary Albedo modality as a structural constraint, the multi-condition RGB + albedo input (bottom) successfully reconstructs the surface normals of the video.}
    \label{fig:discussion}
\end{figure}

\paragraph{Why do we not use vanilla self-attention?} In our \textbf{UniVidX} framework, we employ Cross-Modal Self-Attention (CMSA) instead of the standard vanilla attention. 
While vanilla attention maximally preserves the generative priors of the pre-trained VDM by processing each stream independently, this isolation prevents information exchange among modalities, resulting in weak cross-modal alignment. 
In contrast, our CMSA facilitates interaction by aggregating the keys and values from all modalities to form a shared context, which allows each modality to attend to others, effectively resolving misalignment issues. 
We validate this design using the \textit{UniVid-Intrinsic} instantiation, comparing our full model against the 'w/ Van.' variant equipped with vanilla attention.

As shown in Fig.~\ref{fig:wo_attn}, our model demonstrates strong consistency across all modalities in the text-to-intrinsic task, maintaining precise alignment even in fine-grained details (e.g., the astronaut's suit). 
Conversely, the 'w/ Van.' variant suffers from significant misalignment due to the lack of inter-modal interaction. 
These results empirically verify the effectiveness of our CMSA.

\begin{figure}[t]
    \centering
    \includegraphics[width=\linewidth]{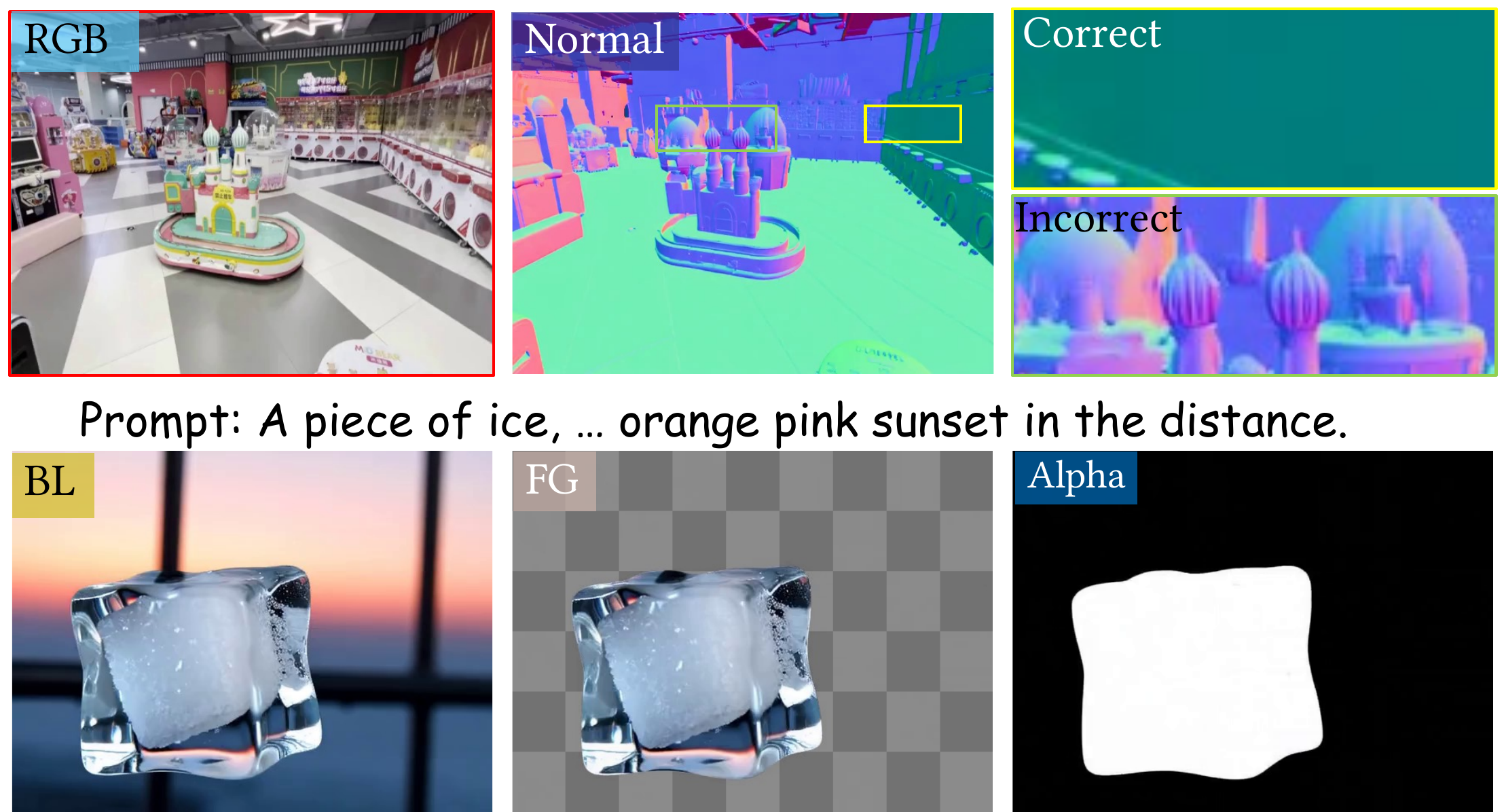}
    \caption{\textbf{Failure cases of our models.} Top row (\textit{UniVid-Intrinsic}): The inverse rendering results given the input RGB (enclosed in a red border). We observe instability in normal estimation for transparent glass surfaces: while it successfully reconstructs the claw machine's glass (highlighted in the yellow box), it fails to capture the geometry of the central glass cover (highlighted in the green box). Bottom row (\textit{UniVid-Alpha}): In the text-to-RGBA task, although the model generates visually plausible BL and FG for the ice cube, the generated alpha matte remains fully opaque (values saturated at 1.0) instead of exhibiting the expected fractional values.}
    \label{fig:failure_case}
\end{figure}
\begin{figure*}
    \centering
    \begin{minipage}{\linewidth}
        \centering
        \includegraphics[width=\linewidth]{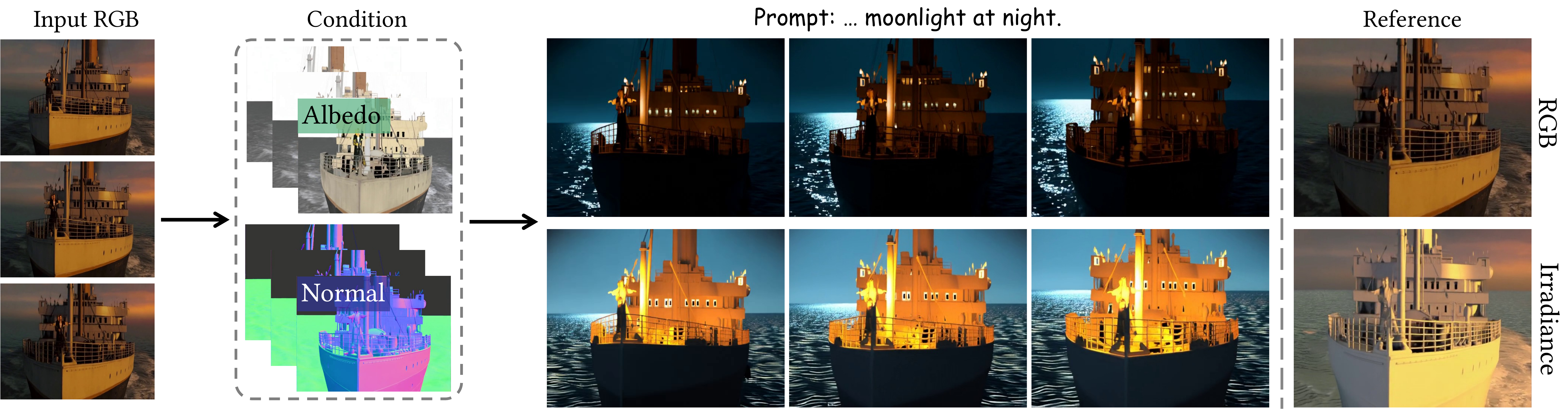}
        \captionof{figure}{\textbf{Application of \textit{UniVid-Intrinsic} --- Video Relighting.} The figure illustrates a two-stage relighting pipeline. First, we perform inverse rendering on the input RGB to get albedo and normal maps. Second, using these intrinsic components as conditions along with a target text prompt, we generate the relighted RGB video and irradiance maps. The reference column displays the original input video and its irradiance from the initial inverse rendering.}
        \label{fig:relighting}
    \end{minipage}
    \begin{minipage}{\linewidth}
        \centering
        \includegraphics[width=\linewidth]{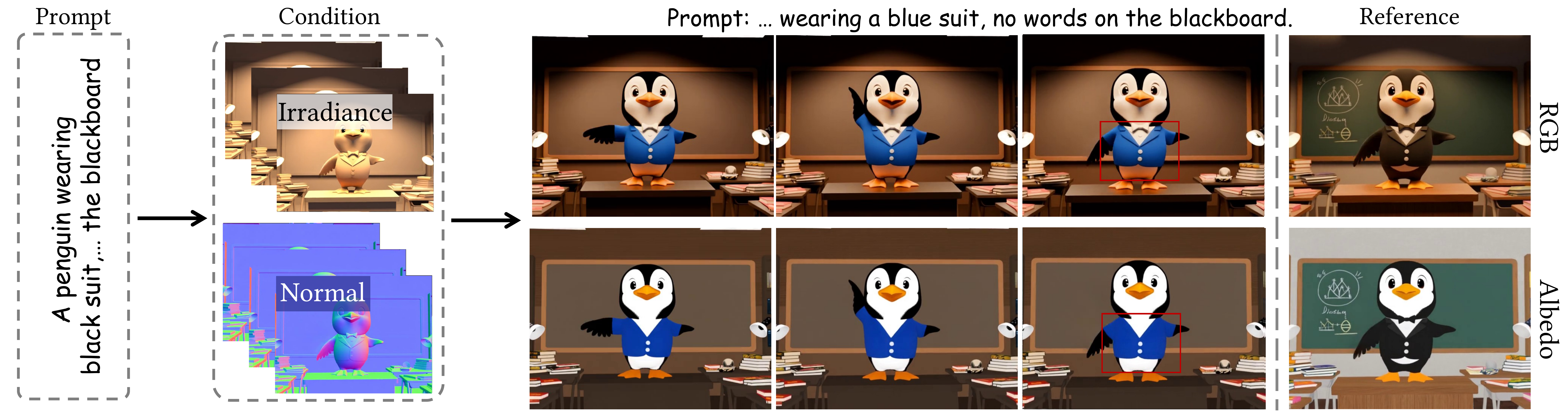}
        \captionof{figure}{\textbf{Application of \textit{UniVid-Intrinsic} --- Text-driven Video Retexturing.} First, we generate the initial RGB and intrinsic maps from a source prompt. Second, we freeze the generated geometry (normal and irradiance) to constrain the structure, while re-synthesizing the RGB and albedo via a target prompt. This pipeline allows for surface appearance control without altering the underlying scene geometry and lighting.}
        \label{fig:retexturing}
    \end{minipage}
    \begin{minipage}{\linewidth}
        \centering
        \includegraphics[width=\linewidth]{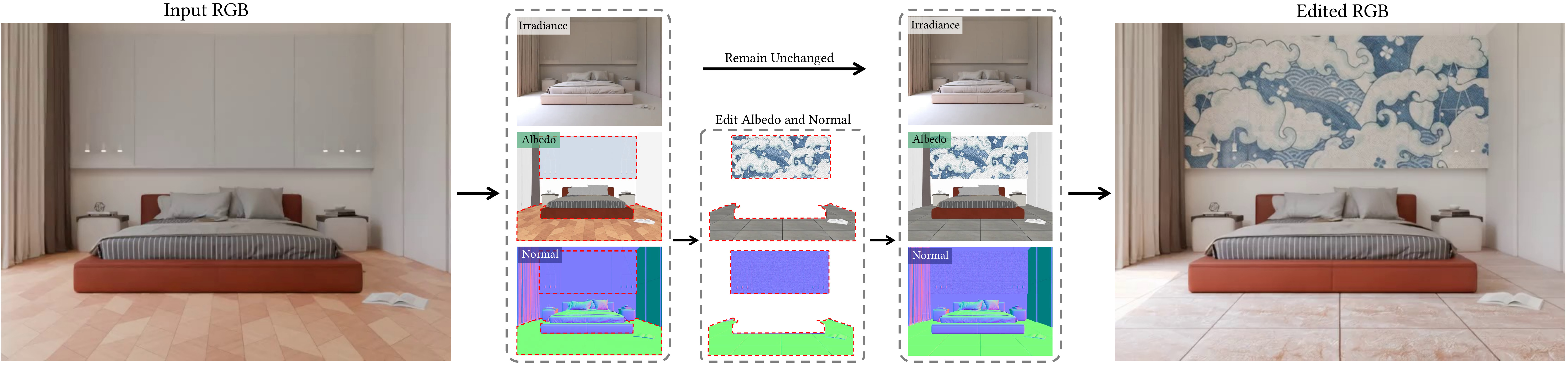}
        \captionof{figure}{\textbf{Application of \textit{UniVid-Intrinsic} --- Material Editing.} First, the input video is decomposed into intrinsic maps. We then manually edit the albedo and normal maps. Finally, taking these edited maps and the original irradiance as conditions, \textit{UniVid-Intrinsic} generates the output with edited materials.} 
        \label{fig:material_edit} 
    \end{minipage}
\end{figure*}

\subsection{The Value of Multi-Condition Perception Paths}
\label{sec:discussion}

Thanks to the flexible generation paradigm of \textbf{UniVidX}, a specific target modality (e.g., normal) can be derived through multiple perception paths (e.g., RGB input; RGB + albedo input).
While standard RGB-based perception (i.e., RGB $\to$ X) generally yields plausible results, we highlight the significant value of multi-condition strategies (i.e., RGB + auxiliary modality $\to$ X) in addressing the inherently ill-posed nature of inverse rendering.
When the RGB input contains ambiguous regions, auxiliary modalities serve as robust semantic cues and structural constraints, guiding the model toward more physically accurate predictions.

A concrete example is illustrated in Fig.~\ref{fig:discussion}.
In the RGB input case, the blurry planet is misinterpreted by the model as empty sky and effectively ignored. 
In contrast, under the RGB + albedo input setting, the additional albedo explicitly signals the presence of the underlying structure, which helps the model accurately recover the surface normals for the planet.

\begin{figure*}
    \centering
    \begin{minipage}{\linewidth}
        \centering
        \includegraphics[width=\linewidth]{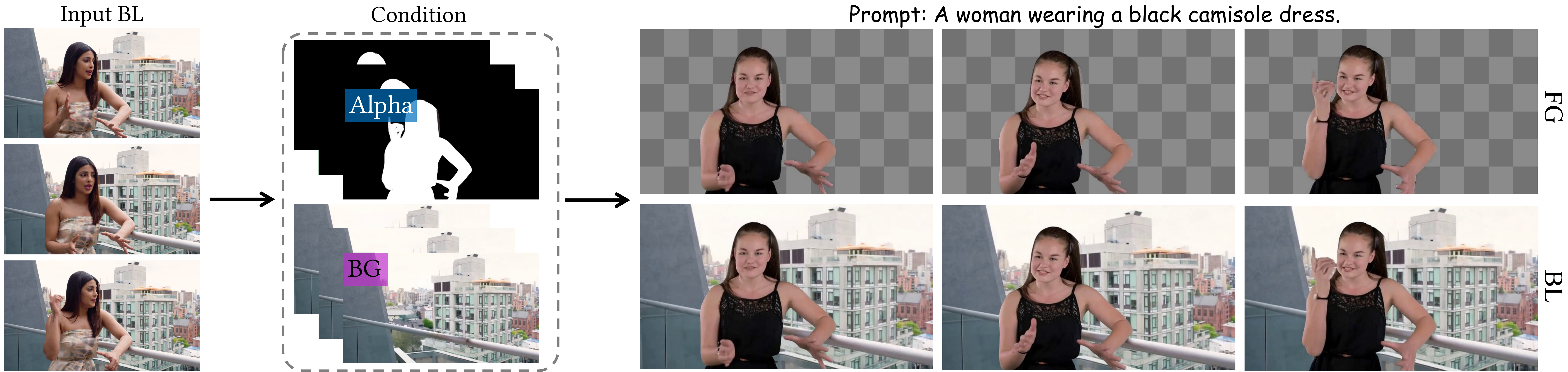}
        \captionof{figure}{\textbf{Application of \textit{UniVid-Alpha} --- Video Inpainting.} First, we decompose the input video into alpha mattes and background components. Second, conditioning on these extracted alpha mattes and background videos, we generate new foreground and blended RGB videos controlled by a text prompt. This allows for precise appearance editing of the subject within the original context.}
        \label{fig:inpainting}
    \end{minipage}
    \begin{minipage}{\linewidth}
        \centering
        \includegraphics[width=\linewidth]{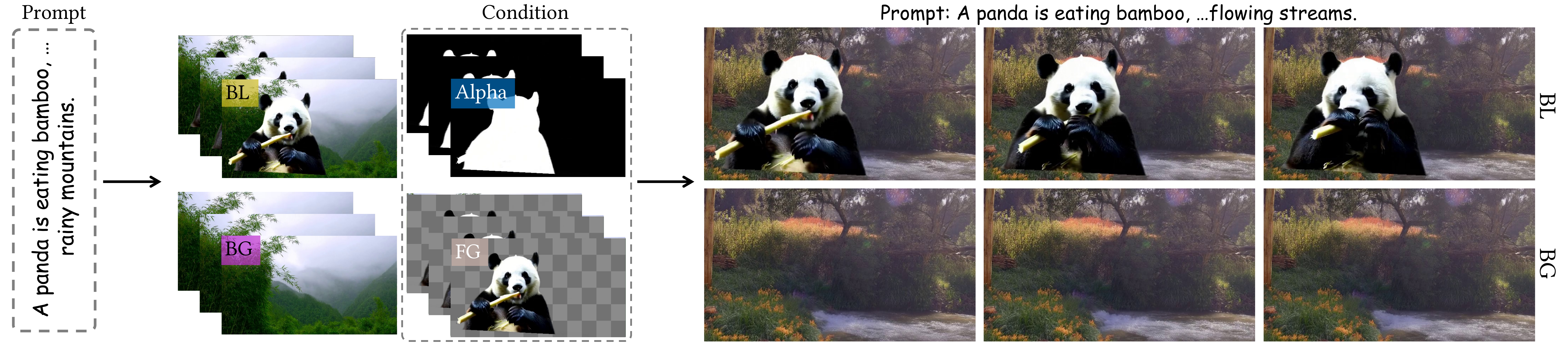}
        \captionof{figure}{\textbf{Application of \textit{UniVid-Alpha} --- Background Replacement.} We first generate the alpha matte and foreground from a text prompt. Then, conditioning on these components and a new background prompt, the model generates the new background and blended RGB output.}
        \label{fig:bg_replacement}
    \end{minipage}
    \begin{minipage}{\linewidth}
        \centering
        \includegraphics[width=\linewidth]{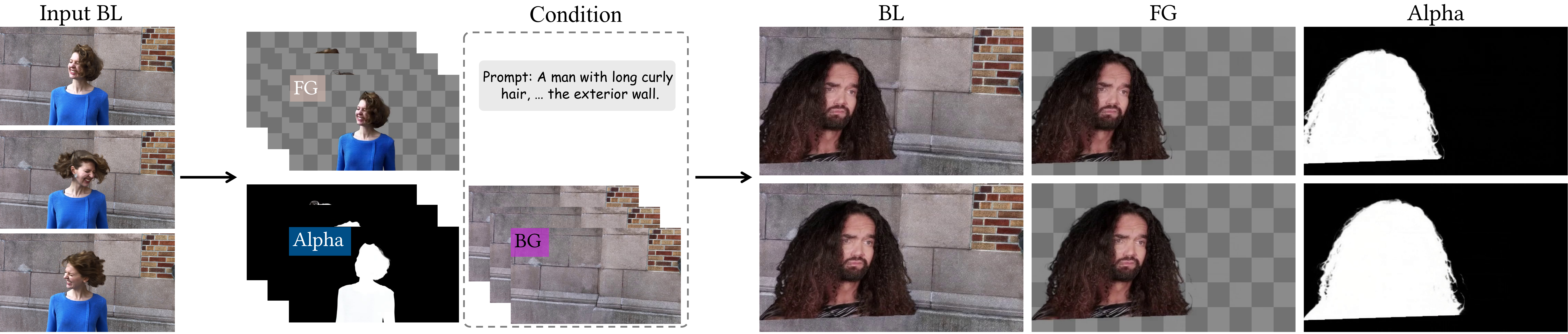}
        \captionof{figure}{\textbf{Application of \textit{UniVid-Alpha} --- Foreground Replacement.} We first extract the background from input video through matting. By conditioning on this background and target foreground prompt, the model synthesizes the corresponding blended RGB, foreground and alpha matte output.}
        \label{fig:fg_replacement}
    \end{minipage}
\end{figure*}

\subsection{Applications}
\label{sec:application}
Benefiting from the versatile generation paradigm of \textbf{UniVidX}, both \textit{UniVid-Intrinsic} and \textit{UniVid-Alpha} support flexible input and output modalities rather than a fixed mapping. This flexibility allows us to creatively combine different tasks within the same model to achieve various downstream graphics applications.

\paragraph{Video Relighting.} As shown in Fig.~\ref{fig:relighting}, we first perform inverse rendering on the input RGB video to obtain intrinsic maps. 
We then select the albedo and normal maps as conditions. 
Combined with a target text prompt, the model generates the relighted RGB video and corresponding irradiance maps. 
Conditioning on albedo and normal ensures that the surface colors and geometric structures remain preserved, allowing only the illumination to be changed.
\paragraph{Text-driven Video Retexturing.} 
Illustrated in Fig.~\ref{fig:retexturing}, we first utilize the model for text-to-intrinsic generation to synthesize a full set of maps. 
We then extract the irradiance and normal maps to serve as conditions. 
By feeding these maps along with a target prompt, we generate the new RGB video and albedo map. The conditioned irradiance ensures consistent lighting, while the normal map preserves the underlying geometry, facilitating surface modification.
\paragraph{Material Editing.} 
As demonstrated in Fig.~\ref{fig:material_edit}, we first decompose the input RGB video into intrinsic components. 
We then manually edit the albedo (to change colors) and normal maps (to modify texture details). 
Finally, taking these edited maps and the original irradiance as conditions, \textit{UniVid-Intrinsic} functions as a forward renderer to generate the final output with updated materials.
\paragraph{Video Inpainting.} As shown in Fig.~\ref{fig:inpainting}, we first decompose the input video into alpha mattes and background components. 
We then condition the model on these extracted alpha mattes and background videos, along with a target text prompt. 
Finally, the model generates new foreground content and the corresponding blended RGB video. 
This process allows for precise appearance editing of the subject while strictly preserving the original context defined by the background and alpha boundaries.
\paragraph{Background Replacement.} 
Illustrated in Fig.~\ref{fig:bg_replacement}, we first generate the alpha matte and foreground from a source text prompt. 
Subsequently, by conditioning on these generated components along with a prompt describing the replacement background, we synthesize the new background layer and the final blended RGB video.
\paragraph{Foreground Replacement.} 
As shown in Fig.~\ref{fig:fg_replacement}, we first extract the background from an input video through video matting, then utilize this background and a target prompt describing the desired subject as conditions. 
Finally, the model jointly generates the corresponding blended RGB video, the new foreground, and its alpha matte, effectively placing a new subject into the existing scene.

\subsection{Limitations and Failure Analysis}
\label{sec:limitations}

\noindent \quad \textit{Two models.} 
Due to the lack of training data jointly annotated with both intrinsic labels and alpha labels, the intrinsic-related and alpha-related capabilities are currently instantiated separately in \textit{UniVid-Intrinsic} and \textit{UniVid-Alpha}. We believe that, if such jointly annotated data become available, these two capabilities can be further unified into a single model within our framework.

\noindent \quad \textit{Computational Constraints.} Despite employing a parameter-efficient tuning strategy (only training LoRAs), the substantial memory footprint of the 14B Wan2.1-T2V backbone necessitates high VRAM usage. 
Consequently, \textbf{UniVidX} is constrained to processing at most 4 modalities, generating videos of up to 21 frames, and operating at a resolution of 480p.
\paragraph{Data Bias and Corner Cases.} We attribute the exceptional data efficiency of our \textbf{UniVidX} to the rich semantic knowledge encapsulated within the pre-trained VDM priors~\cite{tang2023emergent}. 
Conceptually, our fine-tuning process does not learn representations from scratch but rather steers these powerful priors toward the task-specific manifold~\cite{aghajanyan2020intrinsic,lora,ilharco2022editing}. 
However, this strong reliance on priors renders the model susceptible to distribution biases present in the training dataset, leading to suboptimal performance on specific physical corner cases.

A notable example is observed in \textit{UniVid-Intrinsic} when estimating normals for glass surfaces (see Fig.~\ref{fig:failure_case} top row).
Although the input RGB clearly depicts transparent glass in multiple regions, the model exhibits spatially inconsistent behavior: it correctly reconstructs the planar normal of the claw machine’s glass near the right-side wall, yet fails on the central glass cover, where the estimated normals erroneously penetrate the surface to reflect the internal details.
This dichotomy demonstrates that the model indeed possesses the capability to recognize and represent glass materials (as evidenced by the claw machine’s glass case). 
However, it succumbs to the spatial distribution bias of the indoor training dataset \textsc{InteriorVid}, where peripheral regions are typically planar walls and central regions contain complex objects with high-frequency geometry, thus causing the failure in the center glass cover.

A similar phenomenon is observed in \textit{UniVid-Alpha} (see Fig.~\ref{fig:failure_case} bottom row). 
The transparent ice blocks within the generated blended RGB videos correctly refract the background light, demonstrating that the model inherently understands the physical properties of transparent objects. 
However, it fails to predict the corresponding fractional alpha values. 
We attribute this to the label bias in the training data: the human-centric matting dataset VideoMatte240K lacks labels for transparent objects with semi-transparent alpha mattes, thereby leaving the model without the specific knowledge to determine the correct alpha matte for transparent surfaces.

However, these observations are encouraging, suggesting that the VDM backbone already harbors the physical priors to handle such corner cases. Consequently, we believe that these limitations are not structural but data-dependent, and can be effectively resolved by supplementing the training set with targeted samples.

\section{Conclusion}
In this paper, we present \textbf{UniVidX}, a unified framework for versatile multimodal video generation. 
By synergizing Stochastic Condition Masking with Decoupled Gated LoRA, our approach effectively harnesses robust VDM priors, with Cross-Modal Self-Attention ensuring alignment across modalities. 
Validated through \textit{UniVid-Intrinsic} and \textit{UniVid-Alpha}, our approach demonstrates exceptional performance, superior temporal stability, and robust in-the-wild generalization, all achieved with remarkable data efficiency ($<$1k videos). 
By successfully breaking the boundaries of isolated task-specific paradigms, we envision \textbf{UniVidX} as a common recipe for aligned multimodal video modeling, with broader V2V settings left for future work.

\begin{acks}
This work was partially supported by a grant from the NSFC/RGC Collaborative Research Scheme Project No. CRS\_HKUST605/25.
\end{acks}

\bibliographystyle{ACM-Reference-Format}
\bibliography{sample-base}
\clearpage

\end{document}

%% file: tables/text-to-x.tex
\begin{table}[t]
    \centering
  
    \caption{\textbf{Quantitative comparison for text-to-intrinsic and text-to-RGBA generation tasks.} Best results are \textbf{bolded}. "-" indicates that the metric is not applicable (as IntrinsiX and LayerDiffuse generates images)}
    \label{tab:combined_results}
    \resizebox{\linewidth}{!}{
        \begin{tabular}{lcccccccc}
            \toprule
            & \multicolumn{3}{c}{Temporal Flickering} & \multicolumn{5}{c}{User study} \\
            \cmidrule(lr){2-4} \cmidrule(lr){5-9}
            \textit{\textbf{Text-to-Intrinsic}} & RGB~$\uparrow$ & Albedo~$\uparrow$ & Normal~$\uparrow$ & RGB~$\uparrow$ & Albedo~$\uparrow$ & Normal~$\uparrow$ & TA~$\uparrow$ & MC~$\uparrow$ \\
            \midrule
            IntrinsiX~\cite{intrinsix} & - & - & - & 7.82 & 8.44 & 8.12 & 8.65 & 7.02 \\
  
            \textbf{Our UniVid-Intrinsic} & 0.9876 & 0.9885 & 0.9874 & \textbf{9.34} & \textbf{9.23} & \textbf{9.17} & \textbf{9.04} & \textbf{9.29} \\
            
            \midrule[1pt]
            \textit{\textbf{Text-to-RGBA}} & BL~$\uparrow$ & FG~$\uparrow$ & BG~$\uparrow$ & BL~$\uparrow$ & FG~$\uparrow$ & BG~$\uparrow$ & TA~$\uparrow$ & MC~$\uparrow$ \\
            \midrule
            LayerDiffuse~[Zhang et al.~\citeyear{layerdiffuse}] & - & - & - & 9.12 & 8.91 & 8.41 & 8.89 & 8.61 \\
            \textbf{Our UniVid-Alpha} & 0.9912 & 0.9954 & 0.9891 & \textbf{9.30} & \textbf{9.12} & \textbf{9.25} & \textbf{9.04} & \textbf{9.35} \\
            \bottomrule
        \end{tabular}
    }
\end{table}

%% file: tables/inverse_rendering_and_forward_rendering.tex
\begin{table*}[t]
    \centering
    \caption{\textbf{Quantitative comparison of inverse rendering and forward rendering.} Best results are \textbf{bolded} and second best are \underline{underlined}.}
  
    \resizebox{\linewidth}{!}{
        \begin{tabular}{lccccccccccc}
            \toprule
            & \multicolumn{3}{c}{Albedo} & \multicolumn{3}{c}{Irradiance} & \multicolumn{2}{c}{Normal} & \multicolumn{3}{c}{Forward Rendering} \\
            \cmidrule(lr){2-4} \cmidrule(lr){5-7} \cmidrule(lr){8-9} \cmidrule(lr){10-12}
            Methods & PSNR~$\uparrow$ & LPIPS~$\downarrow$ & SSIM~$\uparrow$ & PSNR~$\uparrow$ & LPIPS~$\downarrow$ & SSIM~$\uparrow$ & MAE~$\downarrow$ & $11.25^\circ$~$\uparrow$ & PSNR~$\uparrow$ & LPIPS~$\downarrow$ & SSIM~$\uparrow$ \\
            \midrule
            RGB$\leftrightarrow$X~\cite{rgbx} & 11.64 & 0.3324 & 0.6462 & \underline{11.29} & \underline{0.3734} & \underline{0.7182} & 18.48 & 50.88 & \underline{13.48} & 0.2728 & \underline{0.6842} \\
            Stable Normal~\cite{stablenormal} & - & - & - & - & - & - & 13.68 & 61.23 & - & - & - \\
            Lotus~\cite{lotus} & - & - & - & - & - & - & 14.51 & 58.21 & - & - & - \\
            NormalCrafter~\cite{normalcrafter} & - & - & - & - & - & - & \underline{12.49} & \underline{64.13} & - & - & - \\
            Diffusion Renderer~\cite{diffusionrenderer} & 13.59 & \underline{0.2624} & 0.6817 & - & - & - & 15.76 & 54.42 & 9.87 & 0.2920 & 0.6142 \\
            Ouroboros~\cite{ouroboros} & \underline{14.21} & 0.2639 & \underline{0.7063} & 9.7309 & 0.4560 & 0.6460 & 14.52 & 57.58 & 13.15 &\underline{0.2701} & 0.6700 \\
            \textbf{Our UniVid-Intrinsic} & \textbf{16.89} & \textbf{0.2248} & \textbf{0.7812} & \textbf{13.46} & \textbf{0.3674} & \textbf{0.7895} & \textbf{11.09} & \textbf{70.52} & \textbf{15.31} & \textbf{0.2567} & \textbf{0.7031} \\
            \bottomrule
        \end{tabular}
    }
\label{tab:inverse_rendering_and_forward_rendering}
\end{table*}

%% file: tables/albedo_estimation.tex
\begin{table}[t]
    \centering
    \caption{\textbf{Quantitative results of albedo estimation on the MAW benchmark.} Different cellcolors refer to \colorbox{bestColor}{best}, \colorbox{secondColor}{2nd-best} and \colorbox{thirdColor}{3rd-best}.}
    \label{tab:albedo_estimation}
    \resizebox{0.85\linewidth}{!}{
        {
        \begin{tabular}{lcc}
            \toprule
            Methods & Intensity ($\times 100$) $\downarrow$ & Chromaticity $\downarrow$ \\
            \midrule
            \citet{bell2014intrinsic} & 3.11 & 6.61 \\ 
            \citet{li2018learning}  & 2.71 &  5.15 \\ 
            \citet{sengupta2019neural} & 2.17 & 6.39 \\ 
            \citet{liu2020unsupervised} & 2.62 & 6.00 \\ 
            \citet{li2020inverse} & 1.41 & 5.64 \\ 
            \citet{luo2020niid} & 1.24 & 4.73 \\ 
            \citet{lettry2018unsupervised} & 2.77 & 8.05 \\ 
            \citet{zhu2022learning} & 1.44 & 4.94 \\ 
            \citet{iid} & 1.13 & 5.35 \\ 
            \citet{chen2024intrinsicanything} & 0.98 & 4.12 \\ 
            \citet{careaga2023intrinsic} & 0.57 & 6.56 \\ 
            \citet{colorful} & 0.54 & \best{3.37} \\
            \citet{rgbx} & 0.82 & 3.96 \\
            \citet{diffusionrenderer} & \second{0.46} & \second{3.53} \\
            \citet{ouroboros} & \third{0.48} & 5.47 \\
            \textbf{Our UniVid-Intrinsic} & \best{0.44} & \third{3.60} \\
            \bottomrule
        \end{tabular}
        }
    }
\end{table}

%% file: tables/appendix_sintel.tex
\begin{table}[t]
    \centering
    \caption{\textbf{Quantitative results of normal estimation on the Sintel benchmark.} Different cellcolors refer to \colorbox{bestColor}{best}, \colorbox{secondColor}{2nd-best} and \colorbox{thirdColor}{3rd-best}.}
    \label{tab:sintel}
    \resizebox{\linewidth}{!}{
        \begin{tabular}{lccccccc}
            \toprule
            Methods & \begin{tabular}[c]{@{}c@{}}Training Frames $\downarrow$\end{tabular} & Mean $\downarrow$ & Med $\downarrow$ & $11.25^\circ$ $\uparrow$ & $22.5^\circ$ $\uparrow$ & $30^\circ$ $\uparrow$ & RanK $\downarrow$ \\
            \midrule
            DSINE & 160K & 34.9 & 28.1 & 21.5 & 41.5 & 52.7 & 5.7 \\
            GeoWizard & 280K & 37.6 & 32.0 & 11.7 & 32.8 & 46.8 & 7.8 \\
            GenPercept & \third{90K} & 34.6 & 26.2 & 18.4 & \third{43.8} & 55.8 & 4.4 \\
            Stable-Normal & 250K & 38.8 & 32.7 & 17.9 & 36.1 & 46.6 & 8.0 \\
            Marigold-E2E-FT & \second{59K} & \third{33.5} & 27.0 & 21.5 & 43.0 & 54.3 & 4.6 \\
            Lotus & \second{59K} & \second{32.3} & \second{25.5} & \second{22.4} & \second{44.9} & \third{57.0} & \second{2.2} \\
            NormalCrafter & 860K & \best{30.7} & \best{23.9} & \best{23.5} & \best{47.5} & \best{60.1} & \best{1.0} \\
            Ours & \best{19K} & \third{33.5} & \third{25.8} & \third{21.6} & 43.2 & \second{57.3} & \third{3.1} \\
            \bottomrule
        \end{tabular}
    }
\end{table}

%% file: tables/video_matting.tex
\begin{table}[t]
    \centering
    \caption{\textbf{Quantitative comparison of video matting}. We benchmark our \textit{UniVid-Alpha} against several methods, categorized into Mask-Guided (MG) approaches (top block) and Auxiliary-Free (AF) approaches (bottom block). Best results are \textbf{bolded} and second best are \underline{underlined}.}
    \label{tab:youtubematte}
    \resizebox{\linewidth}{!}{
        \begin{tabular}{lccccc}
            \toprule
            Methods & MAD~$\downarrow$ & MSE~$\downarrow$ & Grad~$\downarrow$ & dtSSD~$\downarrow$ & Conn~$\downarrow$ \\
            \midrule
            AdaM~\cite{lin2023adam} & 4.80 & 0.76 & \underline{2.15} & 1.45 & \underline{0.30} \\
            FTP-VM~\cite{FTP_VM} & 7.45 & 2.14 & 4.76 & 2.07 & 0.31 \\
            MaGGIe~\cite{maggie} & 4.46 & 0.80 & 2.41 & 1.46 & 0.31 \\
            Matanyone~\cite{matanyone} & \underline{4.37} & \underline{0.74} & 2.57 & \underline{1.42} & \textbf{0.26} \\
            \midrule
            RVM~\cite{rvm} & 5.47 & 0.78 & 2.64 & 1.61 & \underline{0.30} \\
            MODNet~\cite{MODNet} & 10.11 & 4.80 & 5.53 & 2.44 & 0.81 \\
            VM-Former~\cite{vmformer} & 6.25 & 1.48 & 3.13 & 2.24 & 0.37 \\
            \textbf{Our UniVid-Alpha} & \textbf{4.24} & \textbf{0.69} & \textbf{1.86} & \textbf{1.39} & 0.52 \\
            \bottomrule
        \end{tabular}
    }
\end{table}

%% file: tables/ablation_wo_gated.tex
\begin{table}[t]
    \centering
    \caption{\textbf{Quantitative ablation on gating design.} We compare the full model against the 'w/o Gated' variant. Best results are \textbf{bolded}.}
    \label{tab:ablation_gated_lora}
    \resizebox{\linewidth}{!}{
        \begin{tabular}{lcccccccc}
            \toprule
            & \multicolumn{3}{c}{Albedo} & \multicolumn{3}{c}{Irradiance} & \multicolumn{2}{c}{Normal} \\
            \cmidrule(lr){2-4} \cmidrule(lr){5-7} \cmidrule(lr){8-9}
            Methods & PSNR~$\uparrow$ & LPIPS~$\downarrow$ & SSIM~$\uparrow$ & PSNR~$\uparrow$ & LPIPS~$\downarrow$ & SSIM~$\uparrow$ & MAE~$\downarrow$ & $11.25^\circ$~$\uparrow$ \\
            \midrule
            w/o Gating & 15.02 & 0.2884 & 0.7112 & 12.04 & 0.4012 & 0.7058 & 13.01 & 59.75 \\
            \textbf{Our UniVid-Intrinsic} & \textbf{16.89} & \textbf{0.2248} & \textbf{0.7812} & \textbf{13.46} & \textbf{0.3674} & \textbf{0.7895} & \textbf{11.09} & \textbf{70.52} \\
            \bottomrule
        \end{tabular}
    }
\end{table}